\documentclass[journal]{IEEEtran}
\usepackage{amsmath,amsfonts}
\usepackage{algorithmic}
\usepackage{algorithm}
\usepackage{array}
\usepackage[caption=false,font=normalsize,labelfont=sf,textfont=sf]{subfig}
\usepackage{textcomp}
\usepackage{stfloats}
\usepackage{url}
\usepackage{verbatim}
\usepackage{graphicx}
\usepackage{cite}
\usepackage{scalerel}
\usepackage{color}
\usepackage{color}
\usepackage[english]{babel}
\usepackage{hyperref}
\usepackage{mathtools}
\usepackage{comment}
\usepackage{multirow}
\title{Semi-Sparsity for Smoothing Filters}

\author{Junqing Huang$^\ast$,
	Haihui Wang$^\ast$,
	Xuechao Wang,
	Michael Ruzhansky,~\IEEEmembership{Member,~IEEE}
	\thanks{Manuscript received 22 April 2022; revised 02 December 2022 and 17 January 2023 accepted 11 February 2023. Date of publication XX, XXXX, XX; date of current version XX, XXXX, XX. This work was supported in part by the FWO Odysseus 1 (Grant number: G.0H94.18N) and Methusalem programme of the Ghent University Special Research Fund (BOF) (Grant number: 01M01021), and also supported in part by the National Science and Technology Major Project, China (Grant numbers: J2019-I-0001-0001 and J2019-I-0019-0018). Michael Ruzhansky is also supported by EPSRC grant EP/R003025/2. (Corresponding author:Michael Ruzhansky.)}
	\thanks{Junqing Huang, Xuechao Wang and Michael Ruzhansky are with the Department of Mathematics: Analysis, Logic and Discrete Mathematics, Ghent University, Ghent 9000, Belgium; Michael, Ruzhansky is also with the School of Mathematical Sciences, Queen Mary University of London, UK. E-mails: \{Junqing.Huang, Xuechao.Wang, Michael.Ruzhansky\}@UGent.be.} 
	\thanks{Haihui Wang is with the School of Mathematical Sciences, Beihang University (BUAA), Beijing 100191, China. E-mail: whhmath@buaa.edu.cn.}
	\thanks{The code is available on \url{https://codeocean.com/capsule/1755106/tree}.}
	\thanks{$^\ast$ The authors contributed equally to this work.}
	\thanks{Digital Object Identifier no. XX.XXXX/TIP.XXXX.XXXXXXX.}}

\begin{document}

\maketitle
	
\IEEEpubid{0000--0000/00\$00.00~\copyright~2023 IEEE}

\begin{abstract}
In this paper, we propose a semi-sparsity smoothing method based on a new sparsity-induced minimization scheme. The model is derived from the observations that semi-sparsity prior knowledge is universally applicable in situations where sparsity is not fully admitted such as in the polynomial-smoothing surfaces. We illustrate that such priors can be identified into a generalized $L_0$-norm minimization problem in higher-order gradient domains, giving rise to a new ``feature-aware'' filter with a powerful simultaneous-fitting ability in both sparse singularities (corners and salient edges) and polynomial-smoothing surfaces. Notice that a direct solver to the proposed model is not available due to the non-convexity and combinatorial nature of $L_0$-norm minimization. Instead, we propose to solve it approximately based on an efficient half-quadratic splitting technique. We demonstrate its versatility and many benefits to a series of signal/image processing and computer vision applications.
\end{abstract}


\IEEEpubidadjcol

\section{Introduction}
\IEEEPARstart{F}{iltering} techniques, especially the feature-preserving algorithms, have been used as basic tools in signal/image processing fields. This favor is largely due to the fact that a variety of natural signals and visual clues (surfaces, depth, color, and lighting, etc.) tend to spatially have piece-wise constant or smoothing property, between which a minority of corner points and edges --- known as sparse singularities or features formed by discontinuous boundaries convey an important proportion of useful information. On the basis of this prior knowledge, many filtering methods are promoted to have ``feature-preserving'' properties to preserve the sparse features while decoupling the local details or unexpected noise. The prevailing use of ``feature-preserving'' filtering methods is also found in many applications, including image denoise\cite{donoho1995noising, perona1990scale}, tone mapping, and high dynamic range (HDR) image compression\cite{farbman2008edge, he2012guided, tomasi1998bilateral}, details enhancement\cite{farbman2008edge, he2012guided, paris2011local, xu2011image}, stylization\cite{winnemoller2012xdog, lu2012combining}, and so on.

In the literature,  it has been witnessed that a number of ``feature-preserving'' filtering methods have been proposed under different research backgrounds. Despite the varying forms and generalizations, they can be broadly divided into two categories: local and global filters. In general, the output of local filters is computed as a weighted sum of local neighbors based on a pre-defined or computed weighted function, which can be found in median filter\cite{weiss2006fast}, bilateral filter\cite{tomasi1998bilateral}, guided filter\cite{he2012guided}, and so on. Due to the ease of implementation, local filters have been widely used in many practical applications. One bottleneck is that they may ignore the global attributes of signal and may cause some unexpected effects, for example, halo artifacts in image detail enhancement\cite{tomasi1998bilateral} and HDR image compression\cite{fattal2002gradient}. On the contrary, many global methods are proposed under an optimization-based framework that finds a globally optimal solution for signal/image filtering problems. The classical approaches, including total variation (TV)\cite{rudin1992nonlinear}, weighted least squares (WLS) filter\cite{farbman2008edge, min2014fast} and $L_0$-norm gradient minimization\cite{xu2011image}, to some extent, overcome the defects of local methods with better filtering performance. The success is, however, achieved at the much-increased cost of solving large-scale linear or no-linear system equations, which is always not negligible in practice even with the recent endeavor in acceleration. Despite the great success, there is still considerable interest to exploit more powerful and efficient filtering methods to achieve more visual-appealing smoothing results for different applications.

\IEEEpubidadjcol

In the paper, we propose a semi-sparse minimization scheme for a type of smoothing filters. This new model is motivated by the sparsity-inducing priors used in recent cutting-edge filtering methods\cite{ono2017l, xu2011image, ye2013sparse}. It turns out that sparsity priors could help to pursue piece-wise constant filtering results and they are particularly suitable for preserving the sparse features (singularities and edges) existing in signals/images. Nevertheless, it worthy notice that sparsity priors may fail in the regions coexisting sparse features and polynomial smoothing surfaces (See Fig. 1). This failure, as interpreted hereafter, is largely due to the fact that sparse priors may be no longer fully admitted in the polynomial-smoothing surfaces, thereby leading to strong stair-case artifacts when imposing sparse regularization in these regions. In contrast, we illustrate that semi-sparsity priors enjoy more favorable properties in simultaneously fitting both sharpening singularities and polynomial-smoothing surfaces. This major difference leads to a new smoothing behavior that helps to alleviate the drawbacks of many existing filtering methods. We will interpret the semi-sparsity model and show the universal applicability of semi-sparsity prior knowledge for natural images. The main contributions of this paper are summarized as follows:

\begin{itemize}
	\setlength{\itemsep}{0pt}
	\setlength{\parsep}{0pt}
	\setlength{\parskip}{0pt}	
	
	\item The semi-sparsity property is explored with a definition under the context of signal/image filtering backgrounds, and we also demonstrate its universal applicability on natural signal/images with a statistic verification.
	
	\item A new semi-sparsity minimization is concisely designed with an $l_0$-norm regularization in higher-order gradient domains, which gives rise to a new filtering method with powerful simultaneous-fitting abilities in both sharpening edges and polynomial-smoothing surfaces.
	
	\item An efficient iterative solver based on a half-quadratic splitting algorithm is introduced to solve the proposed semi-sparsity model in consideration of the non-convexity of $L_0$-norm regularization minimization.
	
	\item A series of experimental results on natural images are also presented to show the favorable properties of the proposed model in comparison of many cutting-edge ``feature-aware'' filtering methods.
	
\end{itemize}

We additionally remark that the simultaneous-fitting ability of the semi-sparsity model makes it possible to remove the local low-amplitude details and preserve the significant but sparse singularities without introducing the notorious stair-case artifacts, especially in polynomial smoothing surfaces.
This contributes to a new and effective filtering tool for a type of signal/image processing tasks, in which traditional methods can not be well-posed both theoretically and practically.

The rest paper is organized as follows. In Sec. \ref{sec:intro}, the related work is discussed in the context of a generalized optimization framework. In Sec. \ref{sec:methodology}, we derive the proposed semi-sparsity model in detail, where the $L_0$ -norm regularization and its limitations are recalled and semi-sparsity prior knowledge is then discussed with the verification on natural images. The semi-sparsity smoothing model and its accelerated solution are presented in Sec. \ref{sec:model}. The experimental result are also presented in Sec. \ref{sec:experiments} with a variety of applications in Sec. \ref{sec:applications}.  We draw our conclusion and further work in Sec. \ref{sec:conclusion}.

\section{Related Work}
\label{sec:intro}

Within the fields of signal/image processing, a number of filtering algorithms have been proposed to reduce noise or perform a signal/image decomposition. Mathematically, many of them can be, explicitly or implicitly, formulated into an optimization-based framework with the form,
\begin{equation}
\begin{aligned}
\mathop{\min}_{u} {{\left\Vert {u}-Af\right\Vert }_2^2}+\lambda\mathcal{R}(u),
\end{aligned}
\label{eq:1}
\end{equation}
where, ${f}$ and ${u}$ are the observed and output signals\footnote{We, throughout, use the symbols ${f}$ and $u$ as functions or operators and the bold ones ${\boldsymbol{f}}$ and $\boldsymbol{u}$ as their discrete counterparts.}, the weight $A$ specifies a spatially-varying confidence map of smoothness of $f$, $\mathcal{R}(u)$ is known as the regularization term; and ${\lambda}>0$ specifies a trade-off between two terms. We review the related work and show the internal relationships based on Eq. \ref{eq:1} despite the varying forms and generalizations.

Firstly, we interpret a type of local filters with only ${A}$ is considered in Eq. \ref{eq:1}, where $\mathcal{R}(u)$ is ignored, or set ${\lambda}=0$ equivalently. As interpreted in\cite{milanfar2012tour}, a general construction of $A$ begins with specifying a kernel function $K$, for example, a Gaussian filter kernel $K$ with the $i$-th element, 
\begin{equation}
\begin{aligned}
{\boldsymbol{K}_{i,j}={\text{exp}( - {(\boldsymbol{x}_i-\boldsymbol{x}_j)}^T \boldsymbol{Q}_{i,j} {(\boldsymbol{x}_i-\boldsymbol{x}_j)}) }},
\end{aligned}
\label{eq:2}
\end{equation}
where, ${\boldsymbol{Q}}$ is a symmetric and positive definite (SPD) matrix depending on the feature $\boldsymbol{x}$, for example, ${\boldsymbol{x=u}}$, ${\boldsymbol{Q}_{i,j}} = \frac{1}{h^2}\boldsymbol{I}$ (${h}$ is a control parameter related to the variance of noise). In general, it holds $\boldsymbol{A}=\boldsymbol{\Lambda}^{-1}\boldsymbol{K}$ with $\boldsymbol{\Lambda}$ is nontrivial diagonal matrix with diagonal elements $\boldsymbol{\Lambda}_{j,j} = \sum_{i}{\boldsymbol{K}_{i,j}}$, and $\boldsymbol{A}$ is a (row-) stochastic matrix as its rows sum to one.  

In the literature, a variety of local filters can be cast into the above kernel-based context, for example, box filter and median filter \cite{weiss2006fast}, in which each output is given by a mean or median value of local neighbors. In bilateral filter\cite{tomasi1998bilateral}, a weighted sum of local neighbors is treated as the smoothing output by taking both spatial and data-wise distances into account. The guided image filter\cite{he2012guided} takes a similar local strategy based on a ridge regression model to estimate the local filter output  efficiently. Non-local means (NLM) filter\cite{buades2011non} is another case, while the weight is derived from patch-wise data. The non-local idea is also verified in block-matching and 3D filtering (BM3D) \cite{dabov2007image} with high-quality performance. It is also easy to see from Eq. \ref{eq:2} that the kernel ${K}$ decays exponentially, therefore the contribution of the samples far from the center can be neglected in practice. This helps to reduce the computational cost because only a few neighbors are sufficient to provide high-quality approximate results. It is worth noticing that many accelerations\cite{barron2016fast,  elad2005retinex,  gastal2011domain, li2016fast} and extensions of these local filters also benefit from this local property. The interested reader is referred to the surveys\cite{milanfar2012tour, milanfar2013, takeda2007kernel} for more details of different kernel functions.

Secondly, we recall a type of global filter based on Eq. \ref{eq:1}. Differing from local filters, these global filters are highly dependent on the regularization term $\mathcal{R}(u)$ to penalize the strength of smoothness, while the weight $\boldsymbol{A}$ is always reduced into identity matrix ($\boldsymbol{A = I}$) for simplicity. This strategy has been adapted by many existing filtering methods, for example, total variation (TV) regularization\cite{rudin1992nonlinear} and many variants\cite{grasmair2010anisotropic, lefkimmiatis2015structure, xu2012structure}, weighted least squares (WLS) methods\cite{farbman2008edge, min2014fast, kim2017fast}, $L_0$ gradient regularization\cite{nguyen2015fast, ono2017l, xu2011image}, and so on. The difference among them mainly lies in the characteristic of prior knowledge for regularization. For the classical TV-based method\cite{rudin1992nonlinear}, the regularization helps to remove the random noise and streaking artifacts while preserving the singularities (corners and edges). Nevertheless, these TV-based methods tend to produce piece-wise constant results, thereby leading to stair-case artifacts in polynomial-smoothing regions\cite{louchet2011total}. In the case of WLS filter\cite{farbman2008edge}, a weighted $L_{2}$ gradient constraint is exploited to reproduce visual-friendly smoothing results, which is proved to be particularly suitable for image smoothing in the case of halo-free image enhancement. When $\mathcal{R}(u)$ comes to $L_0$ gradient regularization\cite{xu2011image}, it is possible to receive a high-quality piece-wise constant solution for signal/image smoothing due to the excellent approximation ability in the flattening regions. However, it still has the limitation in processing a signal coexisting sparse features and polynomial-smoothing surfaces. 

Recently, many approaches\cite{bredies2010total, knoll2011second, parisotto2020higher} have been proposed to infer higher-order regularization to mitigate the stair-case artifacts in polynomial-smoothing surfaces. It has shown in total generalized variation (TGV)\cite{bredies2010total, knoll2011second} that the problem can be alleviated by higher-order regularity in homogeneous regions while still allowing for discontinuities in the singularities. The idea is subsequently extended to a directional case \cite{parisotto2020higher}, allowing smooth signal/images in an anisotropic fashion. Despite the high-quality performance, these higher-order methods usually need to solve an optimization problem, which is time-consuming in situations of large-scale data samples in practice. Besides, many other filtering methods can be understood under the interpretation of Eq. \ref{eq:1} explicitly or implicitly. The wavelet-based denoising methods\cite{donoho1995noising, cai2012image}, for example, may take into account a sparsity-induced regularization for wavelet coefficients because of the sparsity prior knowledge of signal in wavelet transform domains. Many regularization-based filtering methods can be also reformulated into partial differential equations (PDEs) \cite{perona1990scale} to explore the benefits of the edge-preserving property. Despite the different forms and generalizations, they all strive to explore favorable filtering results --- smoothing data but allowing to preserve the singularities and discontinuities, thereby making it easier to extract or analyze useful information.

\section{Methodology}
\label{sec:methodology}

We briefly review of the sparsity-induced regularization for smoothing filters and then interpret the proposed semi-sparse smoothing model with the verification on natural images.

\subsection{\texorpdfstring{$L_0$}{L0}-norm Regularization for Sparsity}

Mathematically, let $v$ be a vector and $v$ is a sparse signal if the majority elements are zeros, denoted as ${\left\Vert v \right\Vert}_0$, where ${\left\Vert \cdot \right\Vert}_0$ is known as $L_{0}$-norm, denoting the number of its non-zero entries of a vector~\footnote{It is not a true norm in a rigorous sense, but intuitively gives a useful metric for the sparsity of vectors.}. $L_{0}$-norm provides a very simple but easily-grasped notion of the sparsity of signal. In the literature, it has shown that many signal/image processing tasks such as sparse recovering \cite{donoho2003optimally, elad2010sparse}, sparse representation and compressed sensing\cite{donoho2006compressed} can be viewed as finding a sparse solution to under-determined (linear) systems. In man cases, the $L_0$ norm has been used as a favorable and powerful regularization tool in the sparsity-induced models\cite{donoho2003optimally, donoho2006compressed, elad2010sparse} to capture the minority of key features (archetypes) of signals. 

In the context of signal/image filtering technique, it has also proved that $L_0$-norm regularization can be beneficial to preserve the minority singularities and sharp discontinuous features of signals. As suggested in\cite{nguyen2015fast, xu2011image}, it is possible to use the $L_0$-norm to measure the non-zeros elements of the gradient vector of an image, giving a regularization model, 
\begin{equation}
\begin{aligned}
\mathop{\min}_{u} {{\left\Vert {u}-{f}\right\Vert }_2^2}+\lambda{\left\Vert \nabla u \right\Vert }_0,
\end{aligned} 
\end{equation}
where $\nabla$ is a gradient operator and $\lambda$ weights the balance of two terms.
The term ${\left\Vert \nabla u \right\Vert }_0$ counts the non-zero elements of gradient signal, which means output $u$ tends to be piece-wise constant. It turns out that such a regularization faithfully helps to preserve singularities, corners, and sharp edges formed by piece-wise constant regions. The regularization idea that the gradients of signals are usually assumed to be sparse---only a tiny part of the gradient magnitudes is relatively large and the remainder is negligible for approximating to zero, underpins different sparsity-induced filtering models\cite{nguyen2015fast, ono2017l, xu2011image}.

\begin{figure}[t]
\centering
\includegraphics[width=0.5\textwidth]{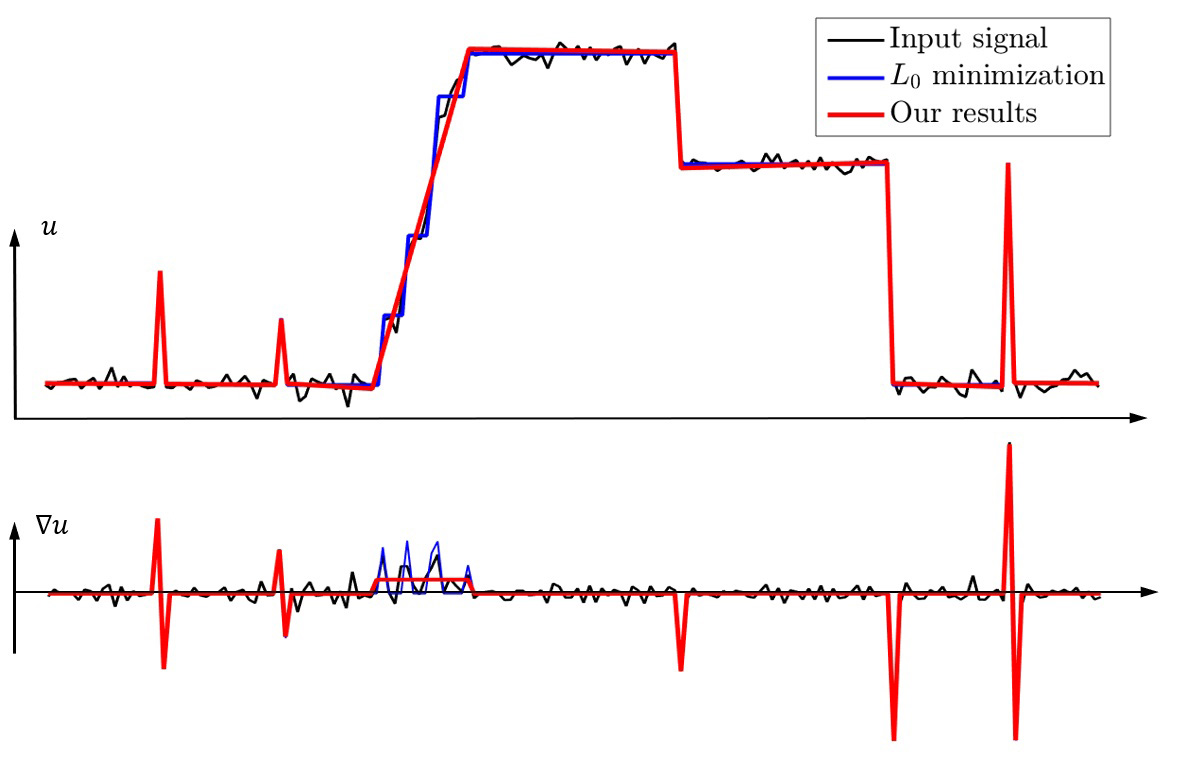}
\caption{An example of 1D signal smoothing, where signal $u$ is composed of the spikes (singularities), sharp edges, and slope line (or, polynomial surface). It is clear that $L_0$ gradient minimization \cite{xu2011image} preserves spikes and sharp edges successfully but produces stair-case artifacts within the slope region (\textcolor{blue}{blue line}). In contrast, our semi-sparsity model attains better fitting (filtering) results in polynomial surfaces with comparable results in both spikes and sharp edges (\textcolor{red}{red line}).}
\label{fig:1}
\end{figure} %

\subsection{Semi-sparsity Priors}

Despite the advantages of $L_0$ gradient minimization\cite{xu2011image}, it may fail in the cases where sparse gradient prior knowledge is not fully admitted, especially in the polynomial smoothing surfaces. A simple example is presented in Fig. \ref{fig:1} for appetite, where a 1D noisy signal $u$ is composed of singularities, sharp edges, and slope lines (or, polynomial surfaces). Clearly, the gradient ${{\nabla} u}$ is not fully sparse within the area of slope region, because the gradient values are relatively larger than zero. As a result, a staircase result (blue) would occur when directly imposing sparse gradient regularization within dense counterparts. This failure is largely due to the fact that sparsity prior knowledge in the gradient domain is no longer fully satisfied, which motivates us to resort to new regularizers for better results --- simultaneously fitting the polynomial surfaces and sparse features (spikes and sharp edges). 

Besides, it is clear in Fig. \ref{fig:1} that the non-zero elements of gradient ${{\nabla} u}$ are not fully sparse but densely distributed within slanted regions. While it is easy to verify that the second-order gradient ${{\Delta} u}$ is sparse in the slanted regions. In this situation, signal $u$ has a semi-sparse property. Without loss of the generality, we say $u$ is a semi-sparse signal if its higher-order gradients ${\nabla}^{n} u$ and  ${\nabla}^{n\!-\!1} u$  satisfy,
\begin{equation}
\small
\begin{aligned}
\begin{cases}
{\left\Vert {\nabla}^{n} u\right\Vert }_0<M,\\
{\left\Vert {\nabla}^{n\!-\!1} u\right\Vert }_0>N,
\end{cases}
\end{aligned}
\label{eq:4}
\end{equation}
where ${\nabla}^{n}$ is the $n$-th (partial) differential operator, and $M, N$ are appropriate natural numbers satisfying ${M \ll N}$. Intuitively, this conclusion can be directly extended into the higher-order ($n \ge 2$) cases. The merit of Eq. \ref{eq:4} is easy to understand, that is, the non-zeros entries of ${\nabla}^{n} u$ is much smaller than that of ${\nabla}^{n\!-\!1} u$, which occurs if and only if ${\nabla}^{n} u$ is much more sparse than ${\nabla}^{n\!-\!1} u$. Taking into account a $n\!-\!1$ degree piece-wise polynomial function for example, it is easy to verify that the $n$-th order gradient ${\nabla}^{n} u$ is sparse, while it not holds for the $k$-th $(k<n)$ order gradient ${\nabla}^{k} u$. This observation motivates us to regularize the higher-order gradients for signal/image restorations. 

\subsection{Verification on Natural Images}
\label{subsec:verification}

In this section, we verify the semi-sparsity prior knowledge of natural images. It is known that the gradient of natural images has a heavy-tailed distribution and can be formulated as a mixture of Gaussian or (hyper) Laplacian models\cite{gong2014image}. We illustrate that the higher-order gradient distributions have similar properties and are possible to be characterized under semi-sparsity prior knowledge. 

Specifically, we take Kodak dataset\cite{kodak} into account, which contains 24 color images with abundant color, structures, and textures information. The normalized distributions of different higher-order gradients ${\nabla}^{k} u$ ($k\!=\!{1,2,3,4,5}$) are presented in Fig. 2. Under mild assumptions\footnote{With a slight abuse of notation, we use $L_0$-norm to measure the sparsity of higher-order gradients for a fair approximation, because these gradients with tiny magnitudes may be zeroed-out during the filtering process.}, it is clear that the higher-order gradient distributions satisfy the following claims:

\begin{itemize}
	\setlength{\itemsep}{2pt}
	\setlength{\parsep}{2pt}
	\setlength{\parskip}{2pt}
	
	\item For natural images $u$,  ${\nabla u}$ admits sparsity criteria: there exists a number $M$, satisfying ${\left\Vert {\nabla}u\right\Vert}_0\!<\!M\!\ll\!N$ ($N$ is the number of pixels), because the majority entries (elements) of ${\nabla} u$ are zero or relatively close to zero. 
	
	\item The higher-order gradient signals of natural images admit semi-sparsity criteria: the higher-order gradients satisfy ${\left\Vert{\nabla}^{i}u\right\Vert}_0\!<\!M_i$, ${\left\Vert{\nabla}^{j}u\right\Vert}_0\!<\!M_j$, where $i\!<\!j$, the positive numbers $M_i\!\ll\!M_j$, because ${\nabla}^{k}u$ tends to be more sparse when the order $k$ goes up.
	
	\item The sparseness of the higher-order gradients of natural images is bounded. Let $M_0\!=\!\max {M_i}$, ${\left\Vert{\nabla}^{k}u\right\Vert}_0\!\le\!M_ 0$ admits for any order $k$, and $M_0$ has a bound, because the gap of sparsity between the gradients ${\nabla}u^{k}$ and ${\nabla}u^{k+1}$ is continuously decreased when the order $k$ increases.
\end{itemize}

\begin{figure}[t]
	\centering
	{\includegraphics[width=0.5\textwidth]{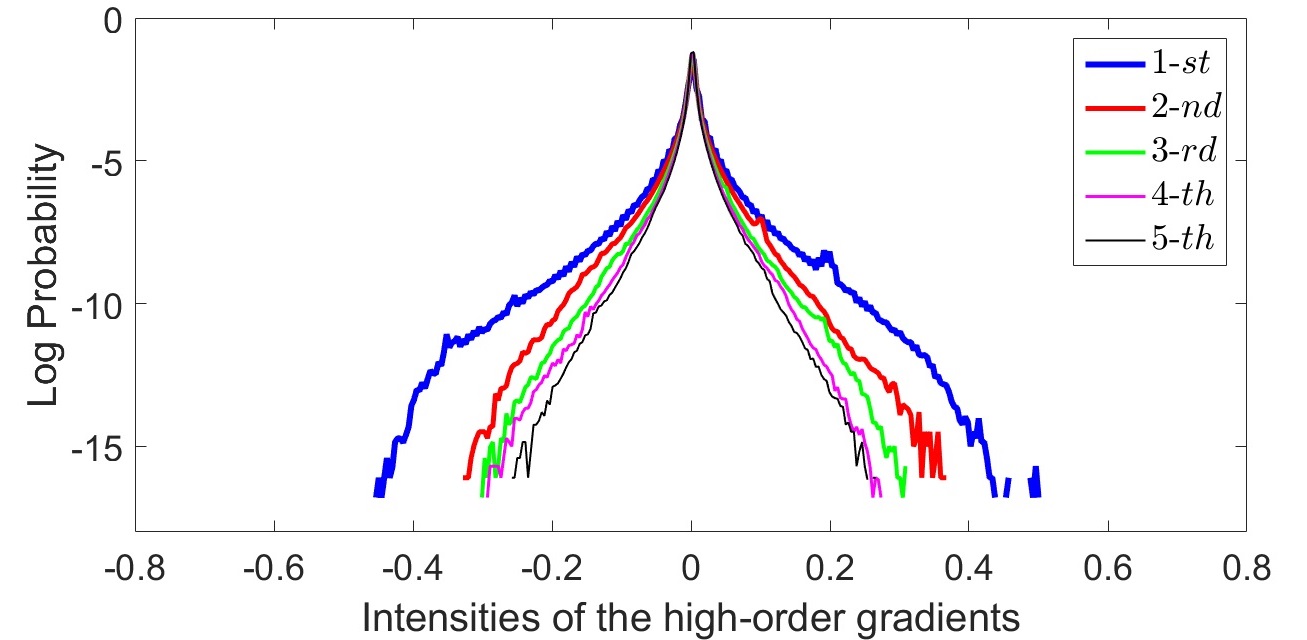}}
	\caption{The normalized distributions of the higher-order ($n\!=\!1\scriptsize{\sim}5$) gradients based on Kodak natural image dataset\cite{kodak}.}
	\label{fig:2}
\end{figure}

The claims can be intuitively verified from the normalized distributions of the higher-order gradients. As shown in Fig. \ref{fig:2}, the first-order gradient of natural images has been well-studied in many sparse-inducing methods\cite{donoho2003optimally, elad2010sparse, xu2011image}. Regarding the higher-order cases, the sparsity of higher-order gradients ${\nabla}u^{k}$ tends to be enhanced as the order $k$ goes up. Moreover, ${\nabla}u^{k}$ also tend to be bounded with the increasing order $k$ in Fig. \ref{fig:2}. This indicates that it is possible to choose an appropriate order depending on the requirements of precision in practical applications. We will illustrate that the above claims could be used for more faithful regularization for our semi-sparsity minimization. 

\section{Semi-sparsity Smoothing Model}
\label{sec:model}

\subsection{Proposed Model}

On the basis of Eq. \ref{eq:4} and verification of natural images, it is naturally to pose a semi-sparse minimization scheme for smoothing filters with the following generalized form, 
\begin{equation}
\small
\begin{aligned}
\mathop{\min}_{u} {{\left\Vert {u}\!-\!{Af}\right\Vert }_2^2}\!+\!\alpha \sum_{k=1}^{n-1}  {\left\Vert {\nabla}^{k}u\!-\!{\nabla}^{k} (Af) \right\Vert }_2^p\!+\!\lambda{\left\Vert {\nabla}^{n} u \right\Vert }_0
\end{aligned}
\label{eq:5}
\end{equation}
where $\alpha$ and $\lambda$ are weights for balance. The first term in Eq. \ref{eq:5} is data fidelity with $A$ adapted from Eq. (1). The second term measures the similarity of $k$-th ($k\!<\!n$) higher-order gradients. The last term indicates that the highest-order gradient ${ {\nabla}^{n} u}$ should be fully sparse under the semi-sparsity assumptions. 

In practice, the choice of $ n $ is determined by the property of signals. For natural images, the claims in \ref{subsec:verification} motivate us to choose $n=2$ as the highest order for regularization. We chose $p=2$ for most filtering cases in view of the non-sparsity property of ${\nabla}^{k}u$ and low computational cost\footnote{It is possible to choose $p$ for other choices, $p=1$, for example, while the numerical solution may be computationally expensive.}. As a result, we have the reduced semi-sparsity model with the form,
\begin{equation}
\small
\begin{aligned}
\mathop{\min}_{u} {\left\Vert {u}\!-\!{f}\right\Vert }_2^2\!+\!\alpha  {\left\Vert {\nabla}u\!-\!{\nabla}f\right\Vert }_2^2\!+\!\lambda{\left\Vert \Delta u \right\Vert}_0
\end{aligned}
\label{eq:6}
\end{equation}
Here, we follow the property of global filters, letting $K$ be Kronecker kernel for simplification.
This choice is based on two-fold benefits: (1) $n=2$ is the simplest case of our semi-sparsity model, and (2) a small $n$ is enough for a wide range of practical tasks with moderate computational cost. 

Mathematically, Eq. \ref{eq:5} (or, \ref{eq:6}) can be viewed as a general extension of the sparsity-induced $L_0$-norm gradient regularization\cite{xu2011image} in the higher-order gradient cases. Despite the minor variants to $L_0$-norm gradient methods\cite{xu2011image, ono2017l}, it has startling different properties compared with many existing filtering methods. On the one hand, it has been shown in many existing filtering methods such as bilateral filter\cite{tomasi1998bilateral}, guided filter\cite{he2012guided}, and TV-based filter\cite{rudin1992nonlinear} that it is possible to guarantee high-quality smoothing results in the polynomial-smoothing areas, but it may inevitably lead to over-smoothing effects around the sharpening edges. On the other hand,  sparsity-induced models\cite{nguyen2015fast, ono2017l, xu2011image} faithfully preserve the sharpening edges without over-smoothing results, but they may produce stair-case artifacts within the polynomial smoothing surfaces. 

In a nutshell, it is difficult for both types of traditional filtering methods to simultaneously fit a signal coexisting the polynomial-smoothing surfaces and sparse singularities features (spikes and edges), since they may either produce over-smoothing results in singularities or introduce stair-case artifacts in the polynomial smoothing surfaces. In contrast, our semi-sparsity minimization model, as demonstrated in the following experimental parts, avoids the dilemma and it is possible to retain comparable edge-preserving properties in strong edges and spikes and more appropriate results in the polynomial-smoothing surfaces, giving arise to a so-called simultaneous-fitting ability in both spikes and polynomial-smoothing regions. This property provides a new paradigm to deal with both cases in high-level fidelity, which contributes its benefits and advantages to many traditional methods. 

\begin{figure*}[t]
	\begin{center}
		\subfloat[Input]
		{\includegraphics[width=0.165\textwidth]{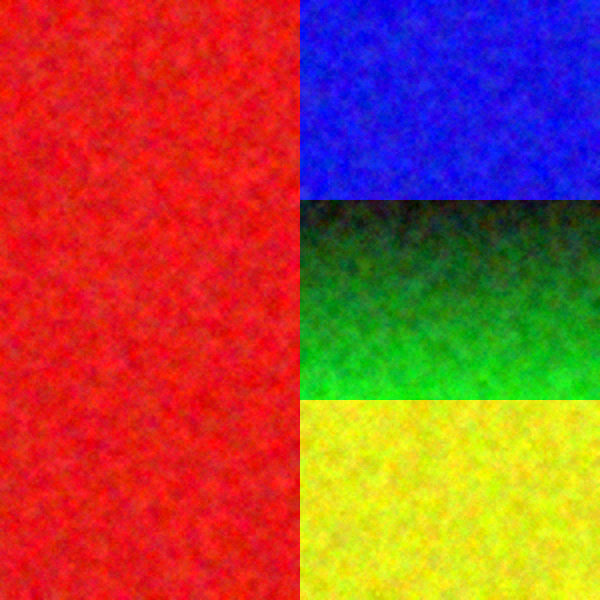}}\hfil
		\subfloat[BF\cite{tomasi1998bilateral}]
		{\includegraphics[width=0.165\textwidth]{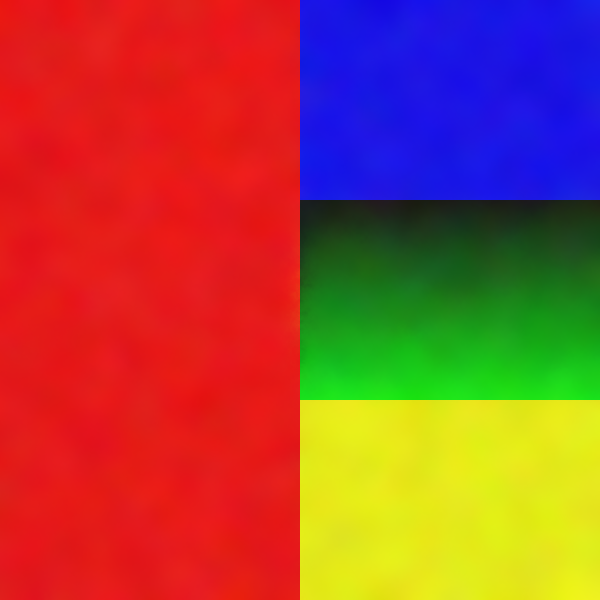}}\hfil
		\subfloat[GF\cite{he2012guided}]
		{\includegraphics[width=0.165\textwidth]{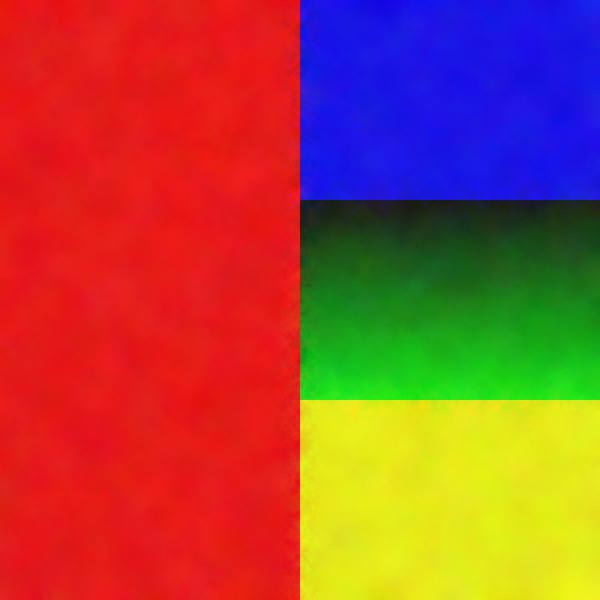}}\hfil
		\subfloat[TV\cite{rudin1992nonlinear}]
		{\includegraphics[width=0.165\textwidth]{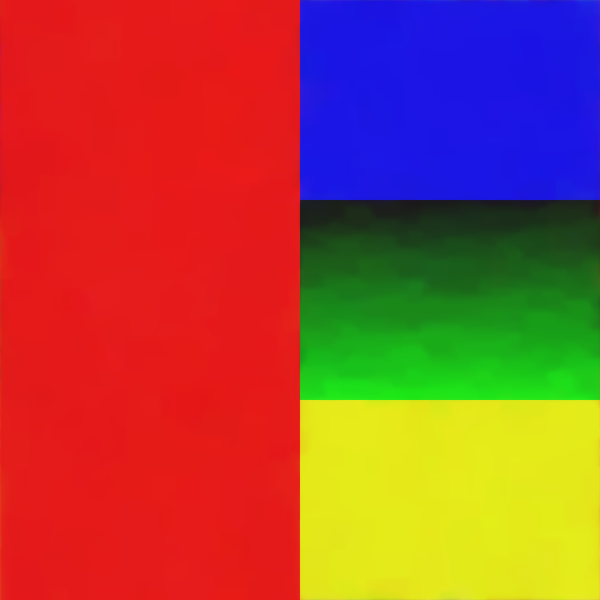}}\hfil
		\subfloat[$L_1$-TV\cite{aujol2006structure}]
		{\includegraphics[width=0.165\textwidth]{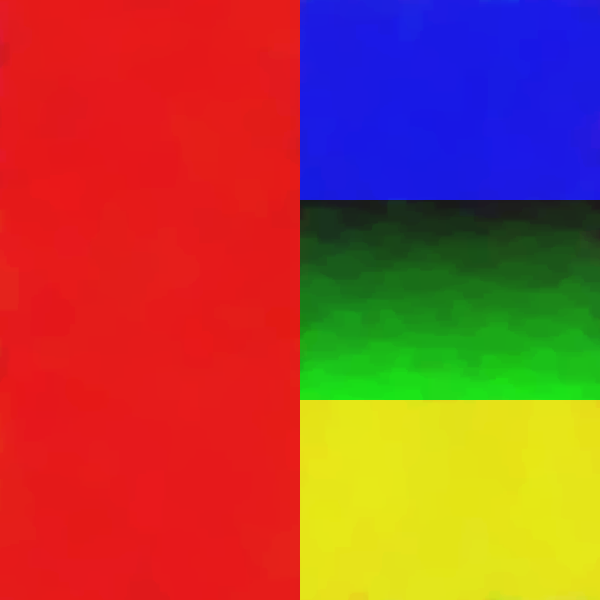}}\hfil
		\subfloat[$L_0$ Gradient\cite{xu2011image}]
		{\includegraphics[width=0.165\textwidth]{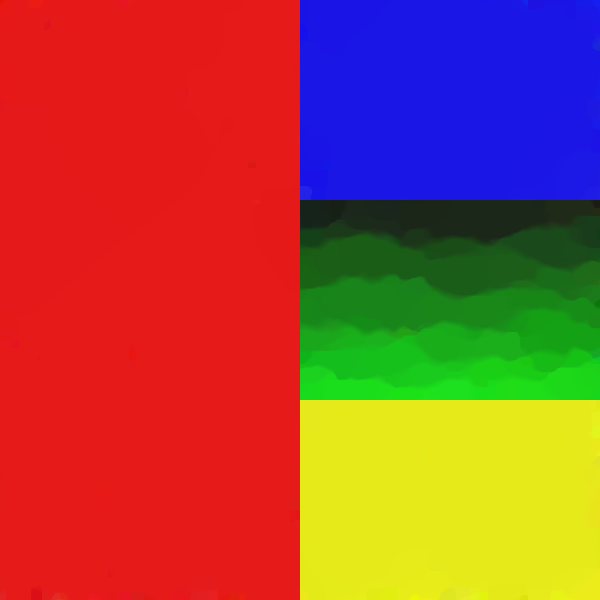}}\hfil
		\subfloat[BM3D\cite{dabov2007image}]
		{\includegraphics[width=0.165\textwidth]{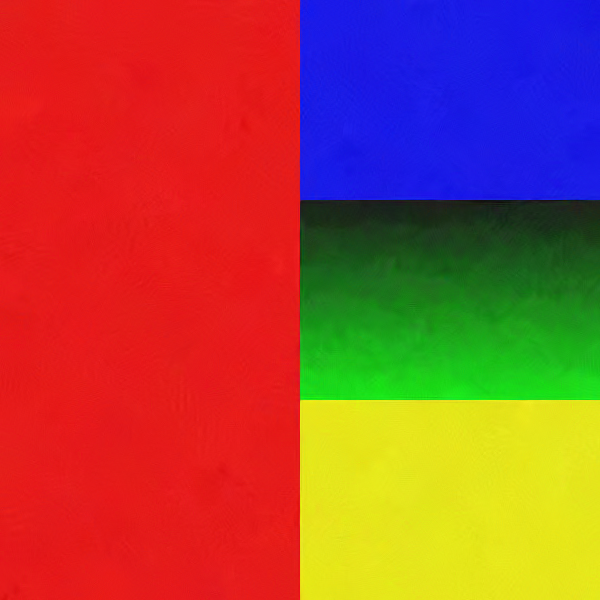}}\hfil
		\subfloat[WLS\cite{farbman2008edge}]
		{\includegraphics[width=0.165\textwidth]{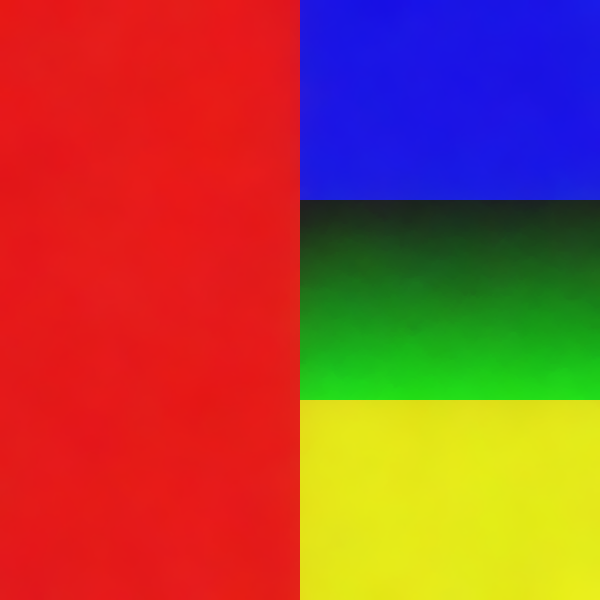}}\hfil
		\subfloat[RTV\cite{xu2012structure}]
		{\includegraphics[width=0.165\textwidth]{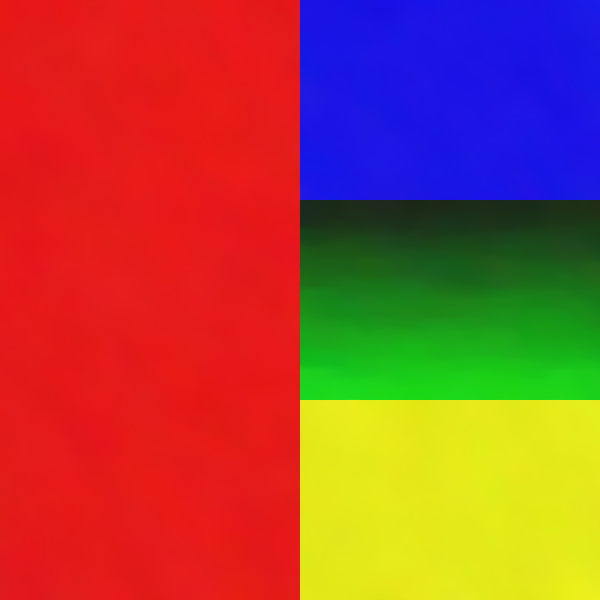}}\hfil
		\subfloat[TGV\cite{knoll2011second}]
		{\includegraphics[width=0.165\textwidth]{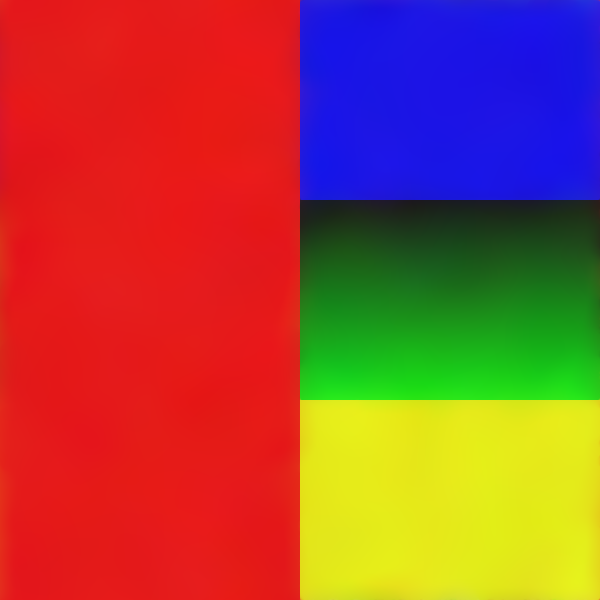}}\hfil
		\subfloat[Ours ($n=2$)]
		{\includegraphics[width=0.165\textwidth]{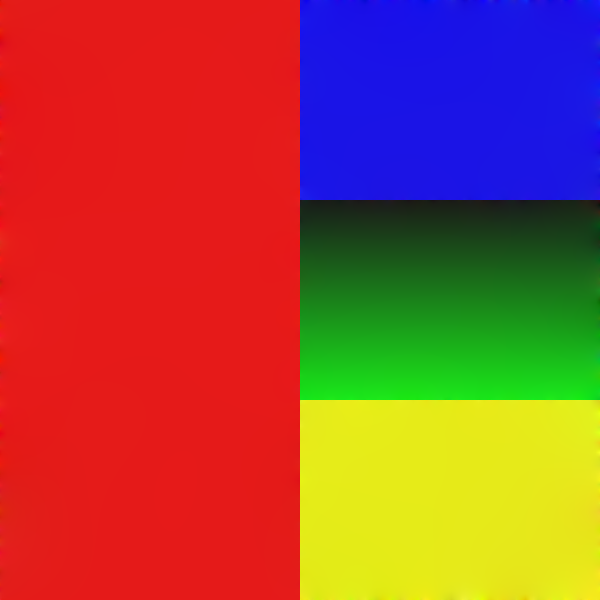}}\hfil
		\subfloat[Ground Truth ]
		{\includegraphics[width=0.165\textwidth]{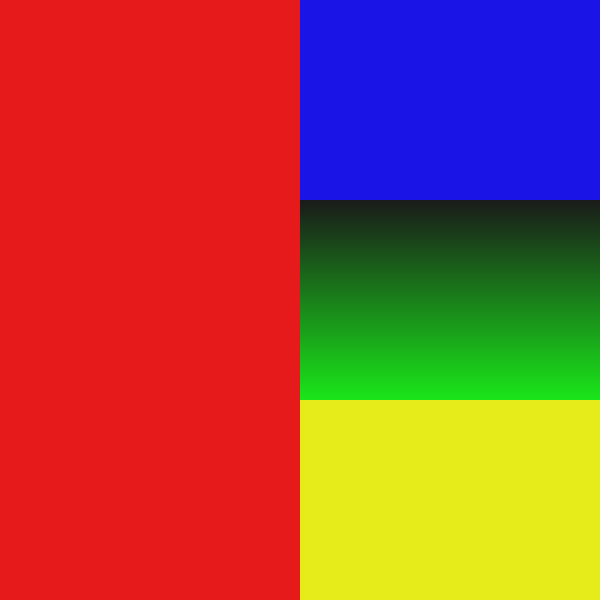}}\hfil
	\end{center}
	\caption{An visual comparison of filtering methods over polynomial-smoothing surfaces. (a) Input,  (b) BF ($w\! = \!25, \delta\! =\![8, 0.25]$)\cite{tomasi1998bilateral}, (c) GF ($w\!=\!25, \epsilon \!=\!0.01^2$)\cite{he2012guided},  (d) TV ($\lambda \!=\! 0.3$)\cite{rudin1992nonlinear}, (e)$L_1$-TV\cite{aujol2006structure}, (f) $L_0$ smoothing ($\lambda \!=\! 0.01, \kappa \!=\!1.05$)\cite{xu2011image}, (g) BM3D \cite{dabov2007image}, (h) WLS ($\alpha \!=\! 0.5, \lambda \!=\! 1.2$)\cite{farbman2008edge},  (i) RTV ($\lambda \!=\! 0.05, \sigma \!=\! 2.0$)\cite{xu2012structure}, (j) TGV ($\lambda \!=\! 0.008, \alpha_0\!=\!0.05, \alpha_1 \!=\! 0.0025$)\cite{knoll2011second},  (k) Ours ($\alpha \!=\! 0.01, \lambda \!=\!0.02$), (l) Ground truth. PSNR: (a)$\scriptsize{\sim}$(k): 27.31, 39.10, 38.40, 39.72, 41.25, 40.50, 41.10, 41.49, 41.63, 42.52, 46.64. We compute PSNR by cropping out 12 boundary pixels for fairness comparison.}
	\label{fig:3}
\end{figure*} %

\subsection{Half-quadratic Solver}
\label{sec:solver}

Despite the simple form of Eq. \ref{eq:5} (or \ref{eq:6}),  a direct solution is not always available due to the non-convexity and combination property of $L_0$-norm. In the literature, such a minimization problem is solved by greedy algorithms\cite{tropp2004greed} to iteratively select the sub-optimal solutions, or reduced with appropriate relaxations\cite{chen2001atomic, wang2008new}. Recently, different efficient algorithms such as half-quadratic (HQ) splitting\cite{allain2006global, wang2008new}, iterative hard-thresholding (IHT)\cite{blumensath2009iterative, sun2017convergence}, and alternating direction method of multipliers (ADMM)\cite{boyd2011distributed, ono2017l} have been proven to be suitable for $L_0$-norm regularization. 

We employ the HQ splitting technique for the semi-sparsity minimization since it is easy to implement and has almost the least computational complexity for large-scale problems. 
The main idea of the HQ splitting algorithm is to introduce an auxiliary variable to split the original problem into sub-problems that can be solved easily and efficiently. For completeness, we briefly introduce the half-quadratic (HQ) splitting technique with a general $L_0$-norm regularized form,
\begin{equation}
\begin{aligned}
\mathop{\min}_{u} \mathcal{F}(u) + \lambda {\left\Vert Hu \right\Vert }_0
\end{aligned}
\label{eq:7}
\end{equation}
where $\mathcal{F}$ is a proper convex function, $H$ is a (differential) operator and the $L_0$-norm regularized term is a non-convex, non-smoothing but a certain separable function.
By introducing an auxiliary variable $w$, it is possible to rewrite the above minimization problem Eq. \ref{eq:7} as,
\begin{equation}
\begin{aligned}
\mathop{\min}_{u} \mathcal{F}(u) + \lambda {c(w)}  + \beta {\left\Vert  Hu \!-\! w \right\Vert }_2^2
\end{aligned}
\label{eq:8}
\end{equation}
where $c(w) = {\left\Vert w \right\Vert}_0$ and ${\left\Vert  Hu \!-\! w \right\Vert }_2^2$ is introduced to measure the similarity of $w$ and $Hu$. The solution of Eq. \ref{eq:8} globally converges to that of Eq. \ref{eq:7} when the parameter $\beta \to \infty$. 

Clearly, the proposed semi-sparsity model is a special case of Eq. \ref{eq:7}, where $\mathcal{F}(u)\! =\! {{\left\Vert {u}\!-\!{f}\right\Vert }_2^2}\!+\!\alpha {\left\Vert {\nabla}(u\!-\!f) \right\Vert }_2^2$ and $H$ is the Laplace operator ${\nabla}$. The optimal solution is then attainable by iteratively solving the following two sub-problems.

\textbf{Sub-problem 1:}  By fixing the variable $w$, the objective function of Eq. \ref{eq:6} reduces to a quadratic function with respect to $u$, giving an equivalent minimization problem,
\begin{equation}
\small
\begin{aligned}
\mathop{\min}_{u} {{\left\Vert {u}-{f}\right\Vert }_2^2}+\alpha {\left\Vert {\nabla}u -{\nabla} f \right\Vert }_2^2 +\beta{\left\Vert \Delta u -w \right\Vert }_2^2.
\end{aligned}
\label{eq:9}
\end{equation}
Clearly, Eq. \ref{eq:9} has a closed-form solution due to the quadratic form w.r.t. $u$. Let $\boldsymbol{D}$ and $\boldsymbol{L}$ be discrete gradient operator and Laplace operator in matrix-form, the optimal solution $u$ is then given by the following linear system:
\begin{equation}
\small
\begin{aligned}
(\boldsymbol{I} + \alpha \boldsymbol{D}^T \boldsymbol{D} + \beta  \boldsymbol{L}^T \boldsymbol{L}) u =  	(\boldsymbol{I} + \alpha \boldsymbol{D}^T \boldsymbol{D}) f + \beta  \boldsymbol{L}^T w,
\end{aligned}
\label{eq:10}
\end{equation}
where $\boldsymbol{I}$ is identity matrix. Since the left-hand side of Eq. \ref{eq:10} is symmetric and semi-positive, many solvers such as Gauss-Seidel and preconditioned conjugate gradients (PCG) methods\cite{saad2003iterative} are applicable for an efficient solution. It is also possible to use fast Fourier transforms (FFTs) for acceleration, which can significantly reduce the computational cost, in particular for a large-scale problem such as millions of variables in the context of image processing.

\textbf{Sub-problem 2:}  By analogy, $w$ is solved with a fixed $u$ and the first two terms in Eq. \ref{eq:6} are constant with respect to $w$, thus it is equivalent to solving the following problem,
\begin{equation}
\small
\begin{aligned}
\mathop{\min}_{w} {\beta} {\left\Vert \Delta u -w \right\Vert }_2^2 + {\lambda}{\left\Vert w \right\Vert}_0
\end{aligned}
\label{eq:11}
\end{equation}
where $c(w)$ counts the number of non-zero elements in $w$. As demonstrated in many existing methods\cite{blumensath2009iterative, wang2008new, xu2011image}, Eq. \ref{eq:11} is separable and can be reduced to  one-dimensional minimization problem --- that is, each variable $w_i$ is estimated individually, giving by the formula,
\begin{equation}
\begin{aligned}
{w_i} =
\begin{cases}
0, &\quad {\left\Vert \Delta u_i \right\Vert }_2^2<\frac{\lambda}{\beta},\\
\Delta u_i, &\quad {\left\Vert \Delta u_i \right\Vert }_2^2 \geq \frac{\lambda}{\beta}.
\end{cases}
\end{aligned}
\label{eq:12}
\end{equation}
It is also known that Eq. \ref{eq:12} is a hard-threshold operator and can be solved efficiently. The HQ splitting algorithm solves two sub-problems iteratively, providing an iterative scheme for the semi-sparsity minimization. Notice that both sub-problems have closed-form solutions in low computational complexity, which makes the $L_0$-norm regularized problem empirically solvable even for a large-scale of variables. The numerical results will demonstrate the effect and efficiency of the HQ splitting scheme.

The above HQ splitting scheme gives rise to an iterative hard-thresholding algorithm, the convergence of which has been extensively analyzed under convex objective function assumptions\cite{allain2006global, wang2008new}. It has shown in \cite{xu2011image} that such a scheme can be directly extended to non-convex $L_0$ gradient regularization for high-quality results. In \cite{blumensath2009iterative}, a similar iterative hard-threshold algorithm is proposed for compressed sensing, where the convergence is built under the restricted isometric property (RIP) and it is also not applicable to our case. It turns out recently that such an iterative thresholding algorithm can be extended to more general non-convex and non-smooth cases, for example, our semi-sparsity minimization, under the Kurdyka–\L{}ojasiewicz property\cite{attouch2013convergence}. The concreted discussion is out of the scope here and the interested reader is referred to the literature\cite{attouch2013convergence, sun2017convergence, zeng2016sparse} for more details. 

\subsection{More Analysis}

\begin{figure*}[!t]
	\begin{center}
		\subfloat[Input]
		{\includegraphics[width=0.19\textwidth]{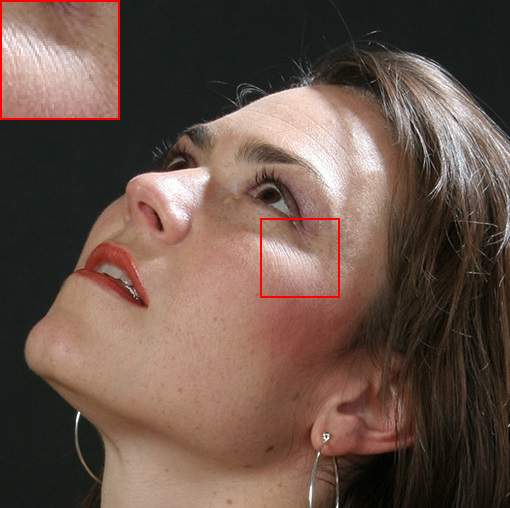}}\hfill
		\subfloat[BF\cite{tomasi1998bilateral}]
		{\includegraphics[width=0.19\textwidth]{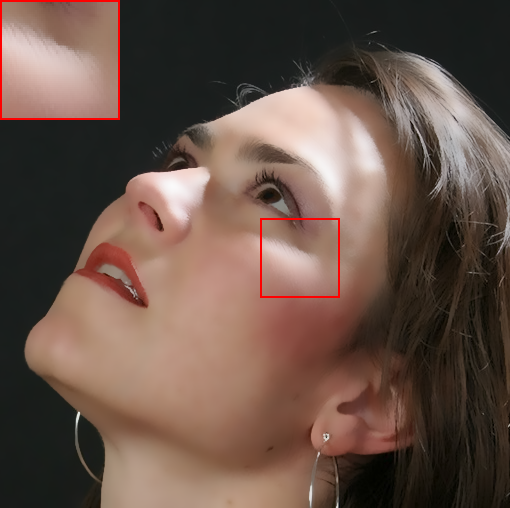}}\hfill
		\subfloat[GF\cite{he2012guided}]
		{\includegraphics[width=0.19\textwidth]{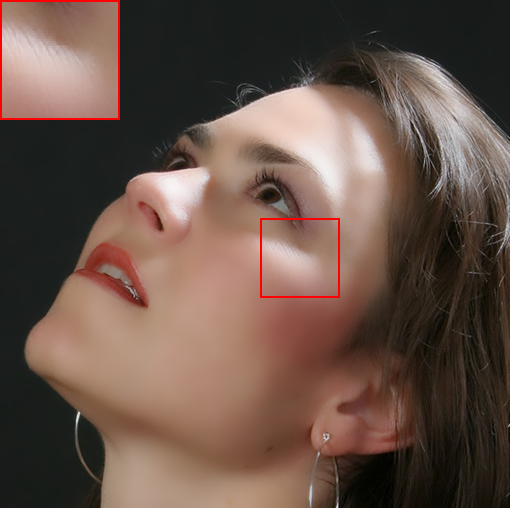}}\hfill
		\subfloat[TV\cite{rudin1992nonlinear}]
		{\includegraphics[width=0.19\textwidth]{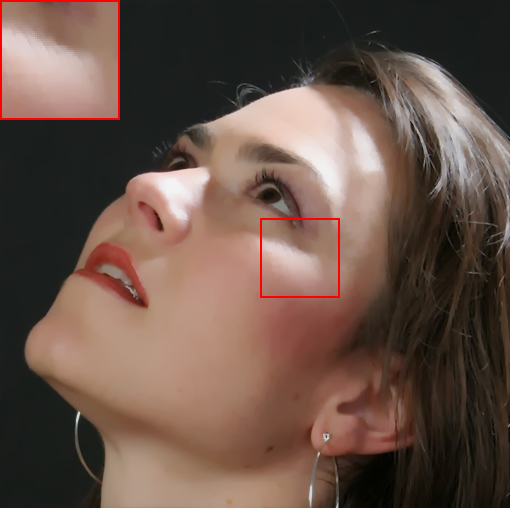}}\hfill
		\subfloat[$L_0$ smoothing\cite{xu2011image}]
		{\includegraphics[width=0.19\textwidth]{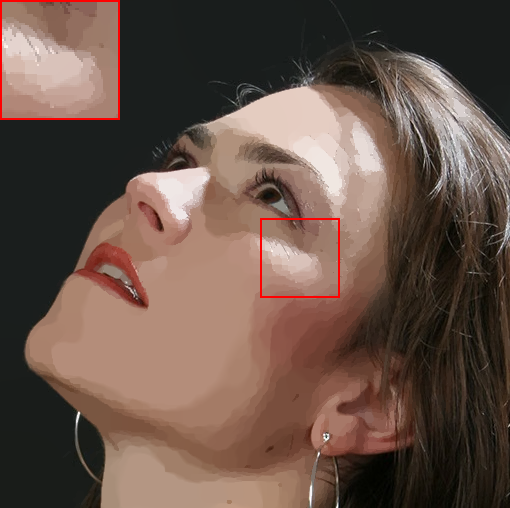}}\hfill
		\subfloat[WLS\cite{farbman2008edge}]
		{\includegraphics[width=0.19\textwidth]{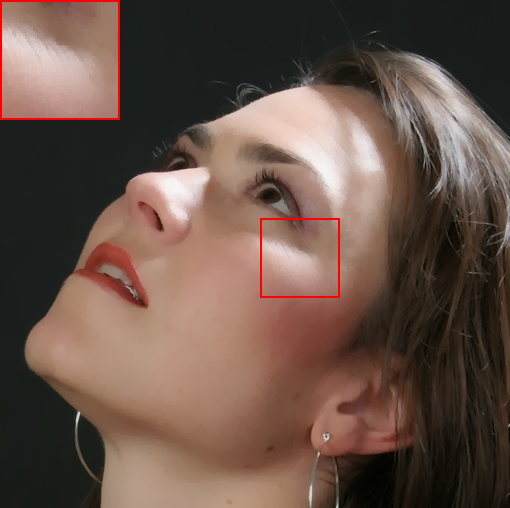}}\hfill
		\subfloat[BM3D\cite{dabov2007image}]
		{\includegraphics[width=0.19\textwidth]{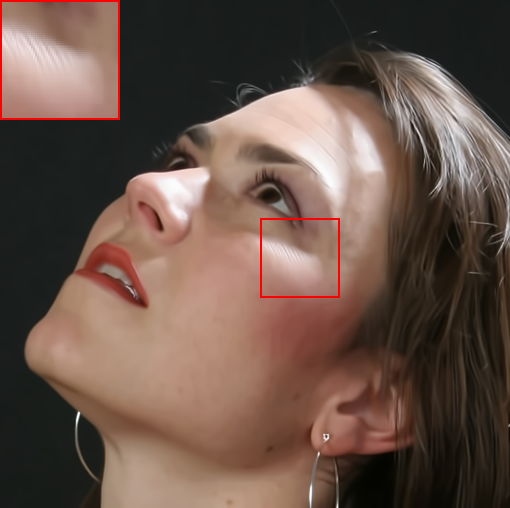}}\hfill
		\subfloat[RTV\cite{xu2012structure}]
		{\includegraphics[width=0.19\textwidth]{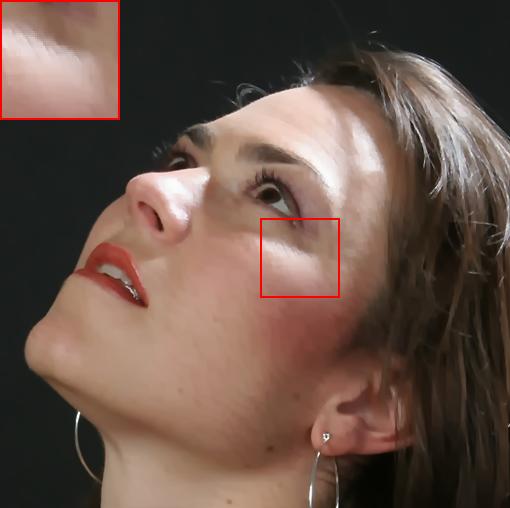}}\hfill
		\subfloat[TGV\cite{knoll2011second}]
		{\includegraphics[width=0.19\textwidth]{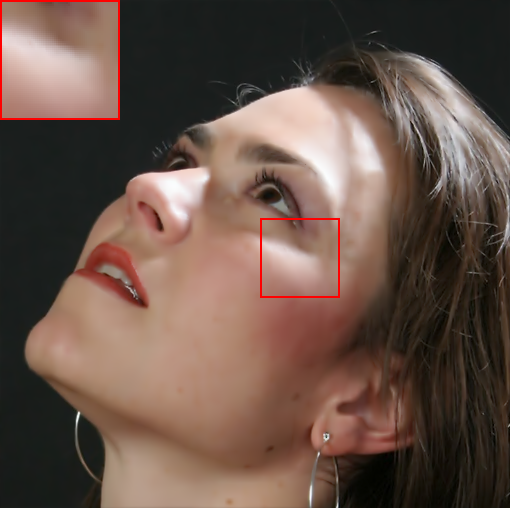}}\hfill
		\subfloat[Ours]
		{\includegraphics[width=0.19\textwidth]{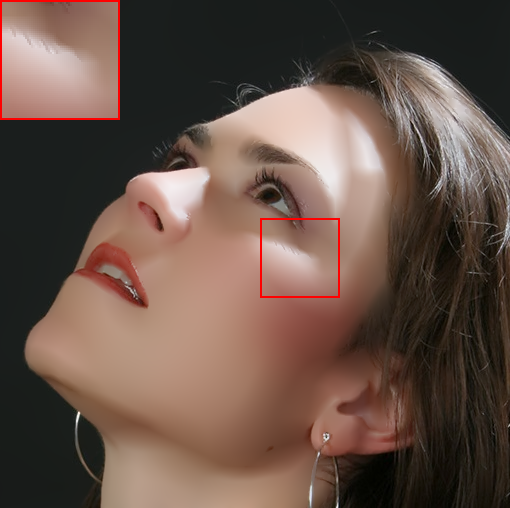}}\hfill
	\end{center}
	\caption{Visual comparison of smoothing results on a face image.  (a) Input, (b) BF ($w\! = \!25, \delta\! =\![8, 0.25]$)\cite{tomasi1998bilateral}, (c) GF ($w\!=\!25, \epsilon \!=\!0.01^2$)\cite{he2012guided},  (d) TV ($\lambda \!=\! 0.2$)\cite{rudin1992nonlinear},  (e) $L_0$ smoothing ($\lambda \!=\! 0.01, \kappa \!=\!1.2$)\cite{xu2011image}, (f) WLS ($\alpha \!=\! 0.5, \lambda \!=\! 1.2$)\cite{farbman2008edge},  (g) BM3D $\delta=30$ \cite{dabov2007image},   (h) RTV ($\lambda \!=\! 0.02, \sigma \!=\! 2.0$)\cite{xu2012structure}, (i) TGV ($\lambda \!=\! 0.01, \alpha_0\!=\!0.002, \alpha_1 \!=\! 0.002$)\cite{knoll2011second},  (j) Ours ($\alpha \!=\! 0.1, \lambda \!=\!0.02$). As suggested in~\cite{weiss2006fast},  the filters are processed in logarithmic intensity for better visual results. MAE: (b)$\scriptsize{\sim}$(j): 0.0115, 0.0117, 0.0118, 0.0116,  0.0117, 0.0119, 0.0117, 0.0116, 0.0117.}
	\label{fig:fig4}
\end{figure*}

\textbf{Parameter Settings:} Notice that the solution of HQ splitting scheme gradually converges to that of Eq. \ref{eq:7} when $\beta$ is large enough. However, it is generally difficult to determine the best value under a given accuracy. We take an adaptive strategy as explained in\cite{wang2008new, he2013mesh}, where $\beta$ is increased by $\kappa$ $(\kappa\!>\!1)$ times in each iteration. A similar configuration is also introduced for $\alpha$, yet, decreasing by $\tau$ $(0 \!\le\! \tau \!<\! 1)$ times in each iteration. The configuration for $\alpha$ is based on the fact that the second term in Eq. \ref{eq:5} plays a similar data-fidelity role in during the smoothing process. 

The parameters $\lambda$, $\alpha$  and $\beta$ control the smoothness of filtering results. We have $\lambda \!\in\! [0.001, 0.1]$, $\alpha \!\in\! [0.01, 10]$ in most cases, and $\beta$ is increased gradually to remove local tiny details until the large-scale or dense features are preserved with little changing. We empirically set $\beta_0 \!=\!\lambda$ and $\beta_{max} \!=\! 1.0e5$ without specification. Additionally, the factors $\kappa$ and $\tau$ are introduced to speed up the convergence of half-quadratic splitting minimization algorithm\cite{he2013mesh, wang2008new, xu2011image}. We set $\kappa \!=\! 1.2$ in most of our experiments without a specification and a smaller $\kappa$ produces more near-ideal results. In contrast, we also set $\tau \!=\! 0.95$ to ensure the smoothing results to be more and more sparse in the highest-order gradient domains. The above configuration maintains a good balance between filtering performance and computational efficiency.

\begin{algorithm}[!t]
	\caption{Semi-Sparse Smoothing Filters}\label{alg:alg1}
	\begin{algorithmic}
		\STATE 
		\STATE {\textbf{Input:}} signal ${f}$, weights ${\lambda, \alpha, \beta}$, parameters $\beta_0$, $\beta_{max}$ \\and rates $\kappa$ and $\tau$;
		\STATE {\textbf{Initialization:}} $u \leftarrow f$, $\beta \leftarrow \beta_{0}$, ${i \leftarrow 0}$;
		\WHILE{$\beta<\beta_{max}$}
		\STATE With $u^i$, solve $w_i$  for Eq. \ref{eq:12};
		\STATE With $w^i$, solve $u_{i+1}$ for  Eq. \ref{eq:10} with FFT acceleration;
		\STATE $\alpha  \leftarrow \tau \alpha$, $\beta \leftarrow \kappa \beta$,  $i\!+\!+$;
		\ENDWHILE
		\STATE {\textbf{Output:}} Smoothing result ${u}$.
	\end{algorithmic}
	\label{alg1}
\end{algorithm}

\textbf{Computational Complexity:} Due to the HQ-splitting scheme, the semi-sparse model can be solved efficiently. As shown in Algorithm~\ref{alg:alg1}, it has two main parts: linear system solver and hard-threshold shrinking in each iteration. The computation is dominated by the FFTs and inverse transforms in the linear system solver at the computational cost of $O(N log(N))$, where $N$ is the total number of pixels in an image. The hard-threshold shrinking operator can be computed in place because of the separable characteristic of the sub-problem. The number of iterations controls the total time of the proposed semi-sparsity algorithm and it roughly needs 15 iterations to produce high-level visual results in most cases. Our Matlab implementation runs on the PC with Intel Core2 Duo CPU 2.13G. It takes roughly 3 seconds to process a $600 \times 400$ resolution color image.

\section{Experimental Results}
\label{sec:experiments} 

In this section, we show the simultaneous-fitting property of the semi-sparse minimization of Eq. \ref{eq:6} and its advantages in producing the ``edge-preserving'' smoothing results. We emphasize the  ``edge-preserving'' property as it is vital for filtering tools to remove the small-scale details but preserve the sharpening variations. The proposed approach is compared with a variety of edge-aware filtering methods, including bilateral filter (BF)\cite{tomasi1998bilateral}, total variation (TV)\cite{rudin1992nonlinear}, $L_1$-TV\cite{aujol2006structure}, guided filter (GF)\cite{he2012guided}, BM3D denoising\cite{dabov2007image}, weighted least square (WLS) filter\cite{farbman2008edge}, $L_0$ gradient minimization \cite{xu2011image}, relative total variation (RTV)\cite{xu2012structure}, total generalization variation (TGV) method\cite{knoll2011second}, and so on. The experimental results are verified on both synthetic and natural images. For the fairness of comparison, the parameters in each filtering algorithm are either configured with a greedy search strategy to give visual-friendly results or fine-tuned to reach a similar level of smoothness, qualified by the mean absolute average (MAE) or peak-signal-noise-ratio (PSNR) metrics.

\textbf{Smoothing Filters:}  We first simulate a $1$-D noisy signal and show the performance for persevering step-wise edges and polynomial smoothing region. As shown in Fig. \ref{fig:1}, the cutting-edge $L_0$ gradient minimization\cite{xu2011image} attains high-quality results in both spikes and step-wise edges, but produces strong stair-case results in polynomial-smoothing regions. As aforementioned, the failure is largely caused by the invalidation of the sparse gradient regularization in the corresponding regions. In contrast, our semi-sparse model gives more accurate fitting results in the slope region with comparable performance in both singularities and sharpening edges. 

\begin{figure*}[!t]
	\begin{center}
		\subfloat[Input]
		{\includegraphics[width = 0.24\textwidth]{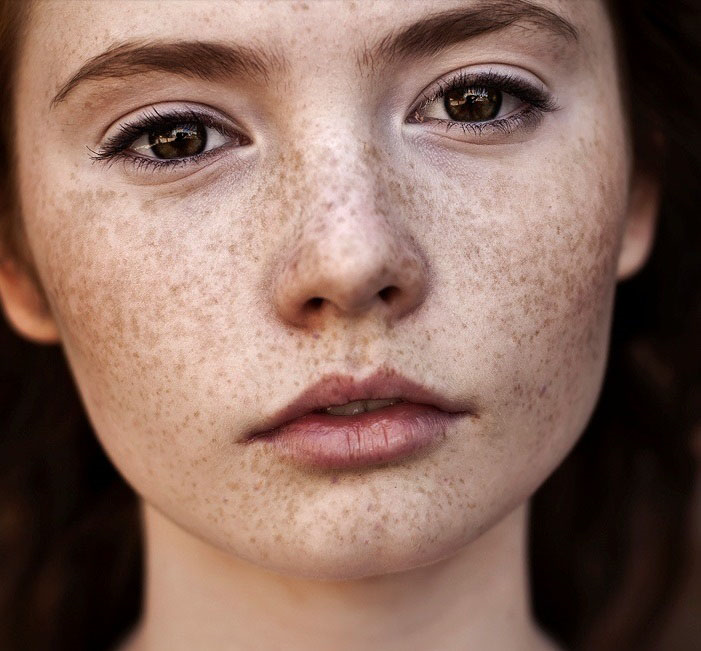}}\hfill
		\subfloat[$\alpha = 10, \lambda =0.002$]
		{\includegraphics[width = 0.24\textwidth]{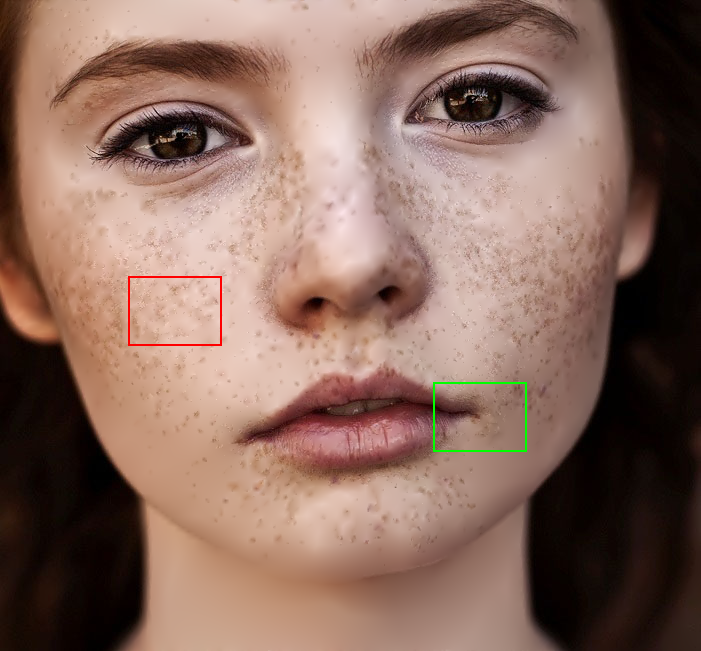}}\hfill
		\subfloat[$\alpha = 10, \lambda =0.01$]
		{\includegraphics[width = 0.24\textwidth]{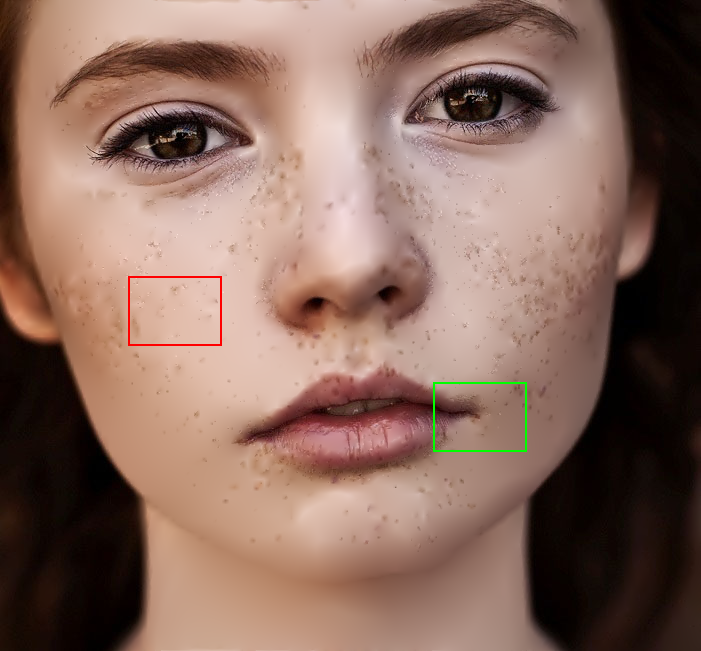}}\hfill
		\subfloat[$\alpha = 10, \lambda =0.1$]
		{\includegraphics[width = 0.24\textwidth]{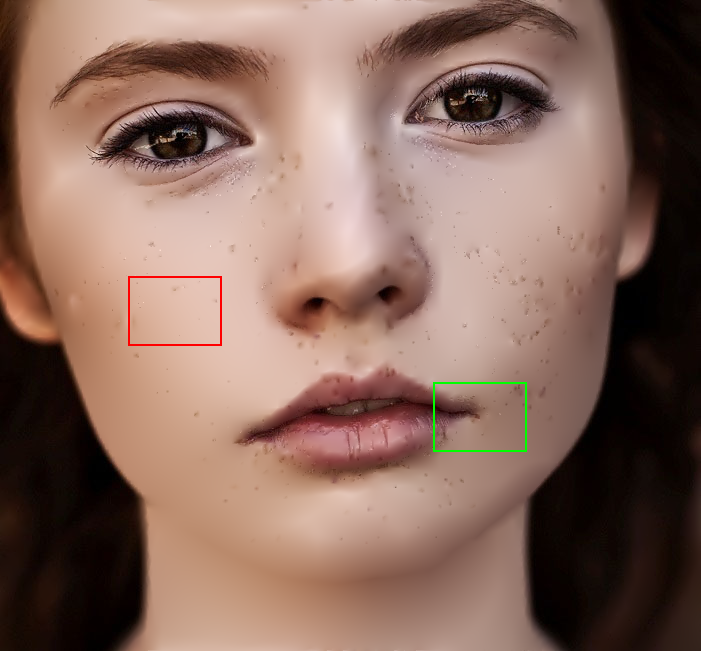}}\hfill
		\hfill
		\subfloat[Close-ups]
		{\includegraphics[width = 0.24\textwidth]{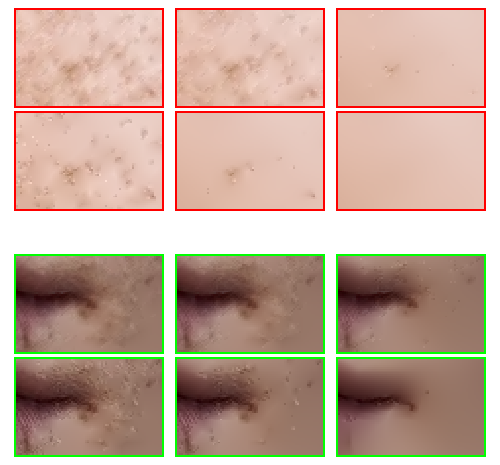}}\hfill
		\subfloat[$\alpha = 0.1, \lambda =0.002$]
		{\includegraphics[width = 0.24\textwidth]{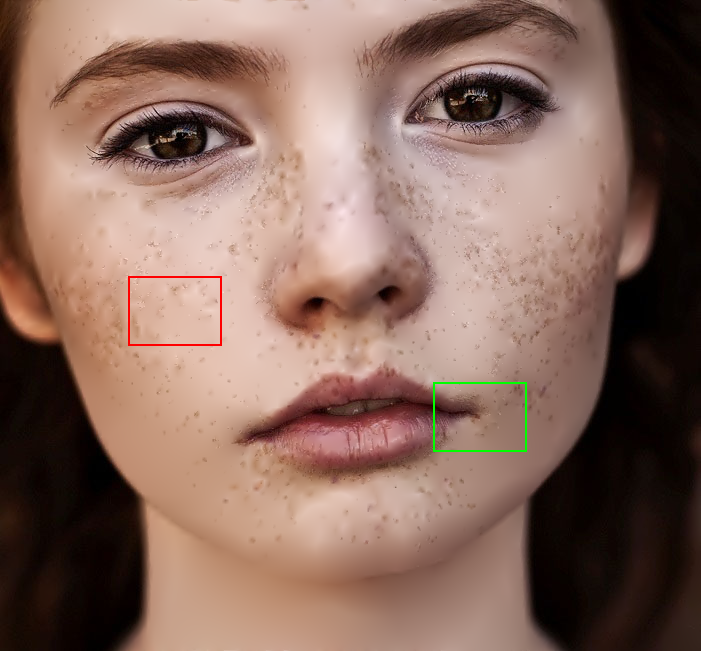}}\hfill
		\subfloat[$\alpha = 0.1, \lambda =0.01$]
		{\includegraphics[width = 0.24\textwidth]{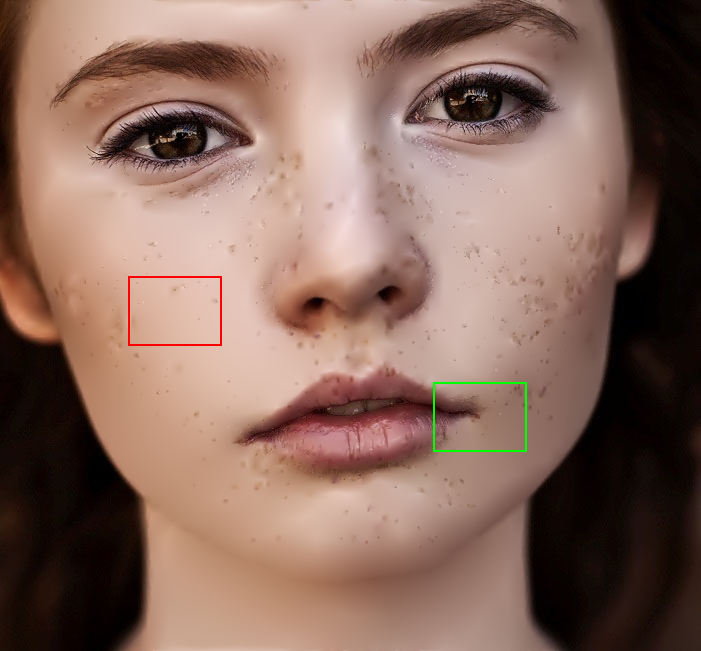}}\hfill
		\subfloat[$\alpha = 0.1, \lambda =0.1$]
		{\includegraphics[width = 0.24\textwidth]{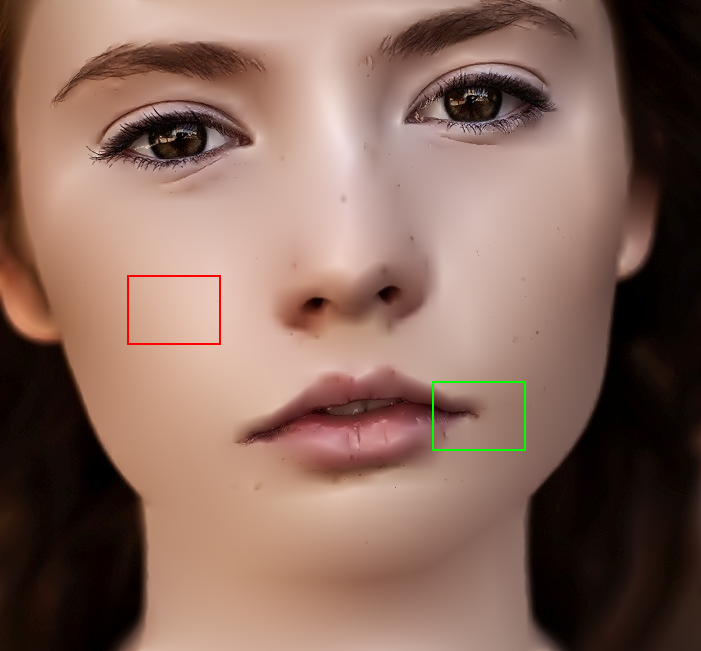}}\hfill
	\end{center}
	\caption{Our semi-sparsity smoothing results of an 2D image with varying parameters $\alpha$ and $\beta$. Note that the filtering algorithm is processed in the logarithmic intensity for better visual effects.} 
	\label{fig:5}
\end{figure*}

To further demonstrate the virtues of our semi-sparse smoothing filter, we compare the results of a synthesized color image that is composed of a piece of polynomial smoothing surface ({\color{green} green region}) and step-wise edges formed by different constant regions.  As shown in Fig. \ref{fig:3},  our method attains similar performance as that in the 1D case. Among them, TV method\cite{rudin1992nonlinear} gives a high-quality fitting result within the interval of slope line, but produces over-smoothing results around the sharpening edge; $L_0$ minimization\cite{xu2011image} preserves the edges but leads to strong stair-case artifacts in the green region and (WLS) filter\cite{farbman2008edge} also gives a high-quality result, yet, with slight deviations in green region. The second-order TGV method\cite{knoll2011second} significantly reduces the stair-case artifacts in the slant surface and also reveals high performance in sharpening edges, while our semi-sparsity exhibits a similar visual result and outperforms around 4 dB improvement under the PSNR metric. Other methods, to some extent, either blur the strong edges or produce piece-wise constant results in the green region. In Fig. 4, we also show the results on natural images. Specifically, we apply the filtering methods on a face image consisting of multiple polynomial surfaces (face) and strong edges (face contour and hair). In this situation, it is usually difficult to produce a visual-pleasant result. The dilemma mainly underpins the fact that both freckles and hair reveal high-frequency characteristics, while the preferred result may require to preserve the details in hair regions but to remove freckles in smoothing face regions. As shown in Fig. \ref{fig:fig4}, most existing ``edge-aware'' filtering methods reveal limited performance in keep the smoothing balance between the high-frequency hairs and smoothing face regions. For example, bilateral filter (BF)\cite{tomasi1998bilateral} and guided filter (GF)\cite{he2012guided} produce acceptable smoothing results in face regions, while they also blur the hair details slightly. BM3D method\cite{dabov2007image} is originally proposed for image denoising and usually reveals high performance for noisy-free results, while it has limited improvement in such cases because of the complexity and diversity of potential details to be filtered out. Notice also that $L_0$ minimization\cite{xu2011image} produces stripe-like results in face regions. In contrast, our semi-sparsity method removes the details in face regions but also retains high-quality edges in the hair region, giving rise to a more promising  result. 

For complementary, we briefly explain the roles of ${\alpha, \beta}$ in controlling the smoothness of the semi-sparse minimization model.  A visual comparison with varying parameters $\alpha$ and $\beta$ is presented in Fig. \ref{fig:5}. In each row, $\alpha$ is fixed and $\lambda$ is gradually increased to give more smoothing results; while, in each column, $\lambda$ is fixed and $\alpha$ is increased to remove more strong local details. As we can see, $\lambda$ plays a crucial role in controlling the smoothness of results and a larger $\lambda$ penalizes to remove more local details and lead to more smoothing results. Over-weighting the second-order term will penalize smoothing regions as well as fine features and therefore over-smooth the details. A similar trend is observed for $\alpha$ but with a decreasing value and the results are not so sensitive to the varying of $\alpha$.  Notice that $\alpha$ is reduced by $\kappa$ time in each iteration, because the penalty for the first-order gradient in Eq. \ref{eq:5} and \ref{eq:6} play a similar role as the data-fidelity term, thus it has a relatively small impact on the global smoothing results but mainly adjusts the local details.
\begin{figure*}[!b]
	\centering
	\subfloat[Input]
	{\includegraphics[width=0.24\textwidth]{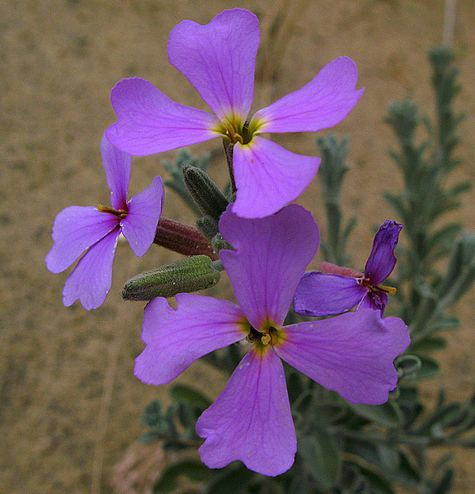}}\hfill
	\subfloat[$1^{st}$-order]
	{\includegraphics[width=0.24\textwidth]{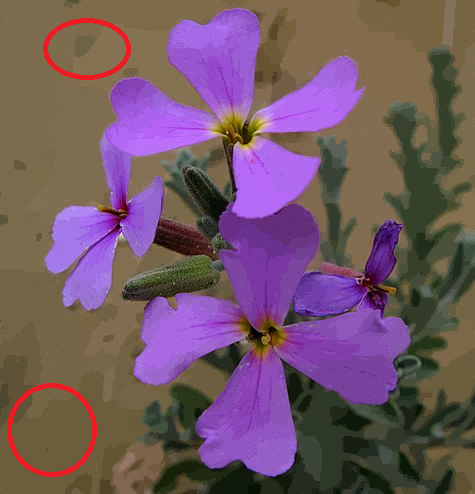}}\hfill
	\subfloat[$2^{nd}$-order]
	{\includegraphics[width=0.24\textwidth]{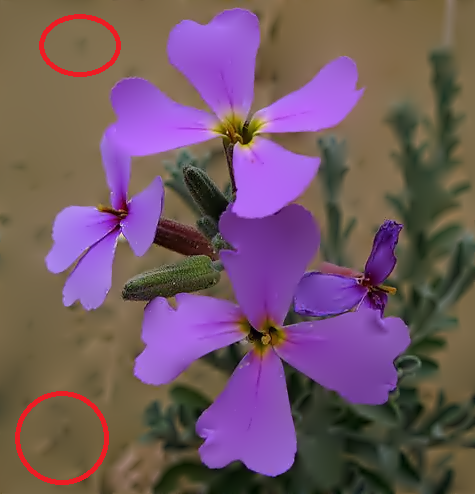}}\hfill
	\subfloat[$3^{rd}$-order]
	{\includegraphics[width=0.24\textwidth]{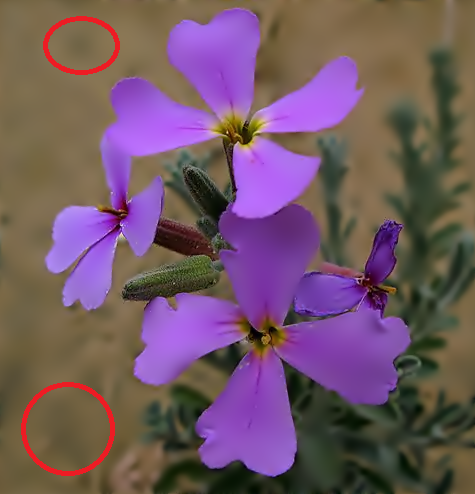}}\hfill
	\caption{Comparison of the smoothing results on a natural image with the sparse regularization from the $1^{st}$-order to $3^{rd}$-order gradient domains, respectively. PSNR: (b)$\scriptsize{\sim}$(d): 33.54, 33.96, 33.16 (Zoom in for better view).}
	\label{fig:6}
\end{figure*}

\textbf{Higher-order Regularization:} The choice of order $n$ in Eq. \ref{eq:5} for sparse regularization is generally determined by the property of signals. For natural images, we interpret that $n\!=\!2$ is usually enough to give favorable results in most cases, although further improvements can be attained with the regularization over higher-order gradient domains. As shown in Fig. \ref{fig:6}, we illustrate this observation by comparing the smoothing results in which the sparsity constraints are imposed from the $1^{st}$ to $3^{rd}$ orders regularizations. As shown in $L_0$-gradient regularization\cite{xu2011image}, the $1^{st}$ order case may introduce strong staircase effects in the flower region and background. Under a similar level of smoothness, the proposed method alleviates the staircase artifacts, and no obvious difference appears except in some tiny regions for the $2^{nd}$ and $3^{rd}$ cases. This result, to some extent, provides validation of the conclusion that the regions having the degree ($n \geq 3$) polynomial surfaces in natural images occupy very tiny regions in natural images, which can be treated as spikes approximately.

This result can be also explained from the higher-order gradient distributions of natural images in Fig. 2, where the sparseness of the higher-order gradient signal is gradually enhanced when the order $n$ increases. Moreover, there exists a bound of the spareness when the order $n$ keeps increasing, as the distributions tend to be more and more similar. On the other hand, it is also clear that the gap of sparseness of the $1^{st}$ and $2^{nd}$ order gradients is much bigger than that of the $2^{nd}$ and $3^{rd}$ order gradients. A similar trend is also observed in the higher-order cases. Accordingly, we claim that the polynomial-smoothing surfaces with degree ($ n \geq 3$) do not frequently occur in natural images, the area of which is relatively smaller than that of the $n=2$ case. Therefore, it is usually enough to impose sparse regularization on second-order gradients for natural images. Another consideration of choosing the $2^{nd}$ order gradient for regularization is to reduce the computation cost in many practical applications.
\begin{figure*}[b]
	\begin{center}
		{\includegraphics[width=0.14\textwidth]{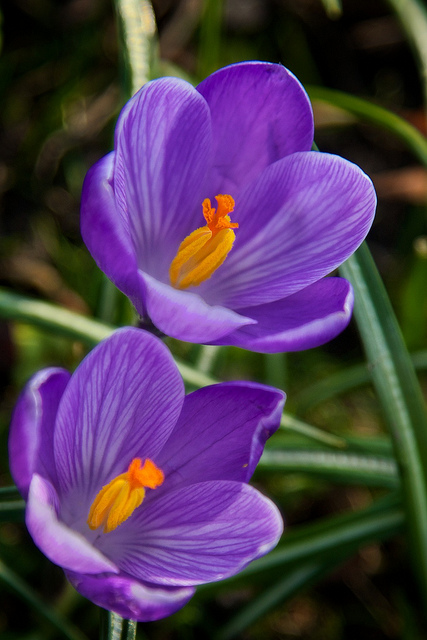}}\hfill
		{\includegraphics[width=0.14\textwidth]{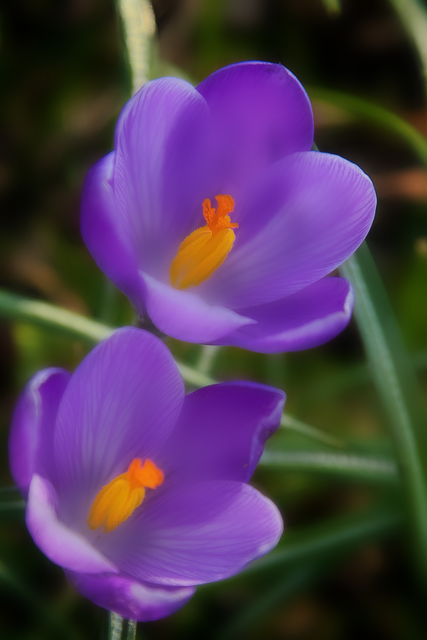}}\hfill
		{\includegraphics[width=0.14\textwidth]{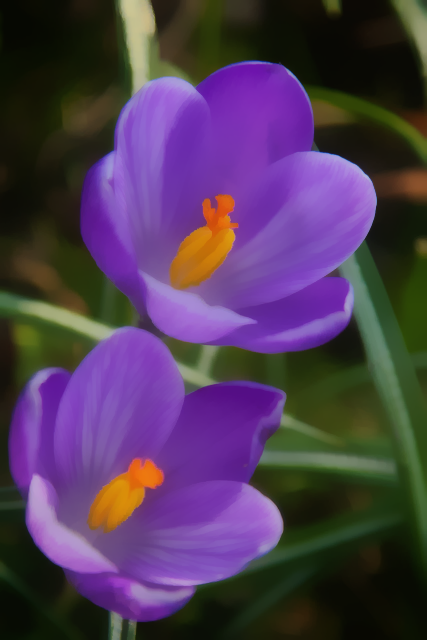}}\hfill
		{\includegraphics[width=0.14\textwidth]{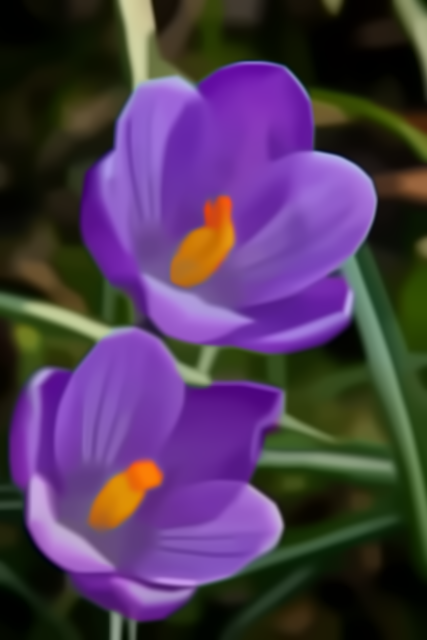}}\hfill
		{\includegraphics[width=0.14\textwidth]{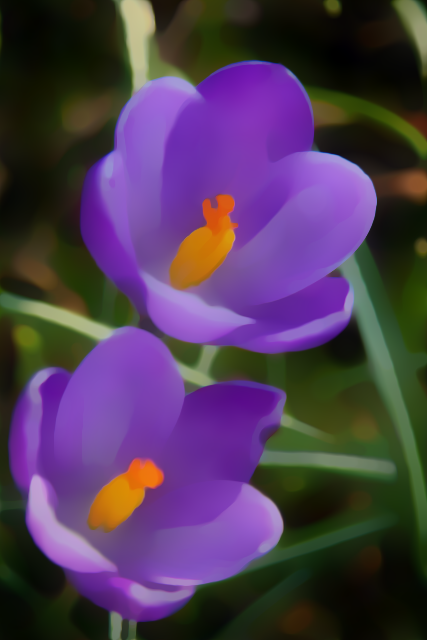}}\hfill
		{\includegraphics[width=0.14\textwidth]{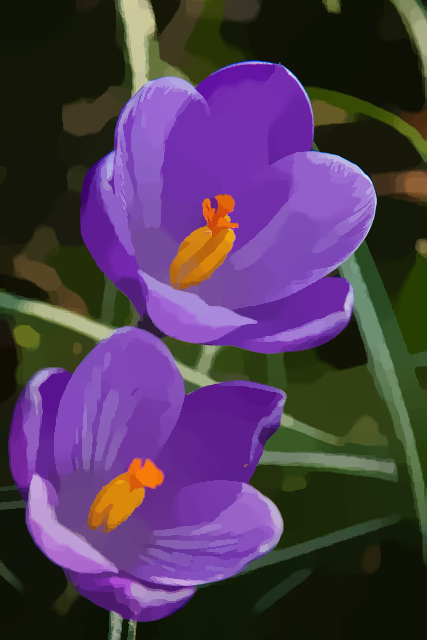}}\hfill
		{\includegraphics[width=0.14\textwidth]{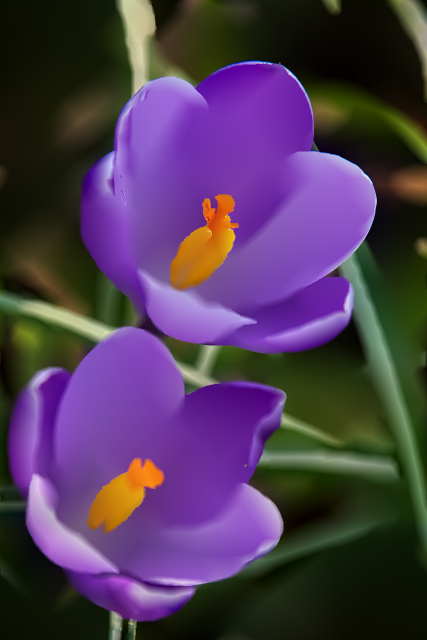}}\hfill
		{\includegraphics[width=0.14\textwidth]{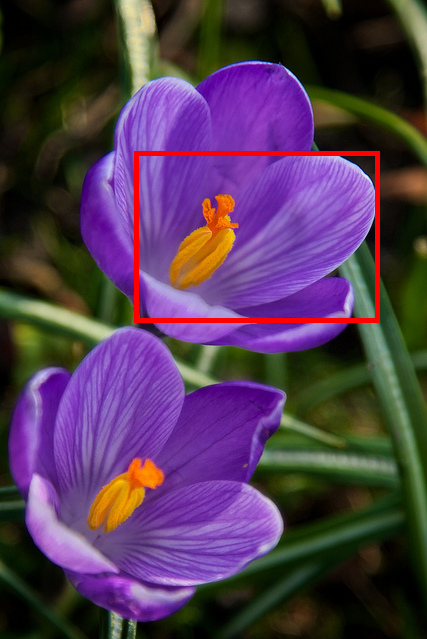}}\hfill
		{\includegraphics[width=0.14\textwidth]{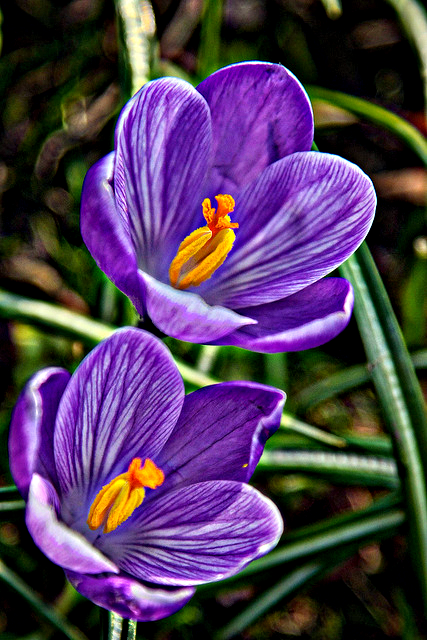}}\hfill
		{\includegraphics[width=0.14\textwidth]{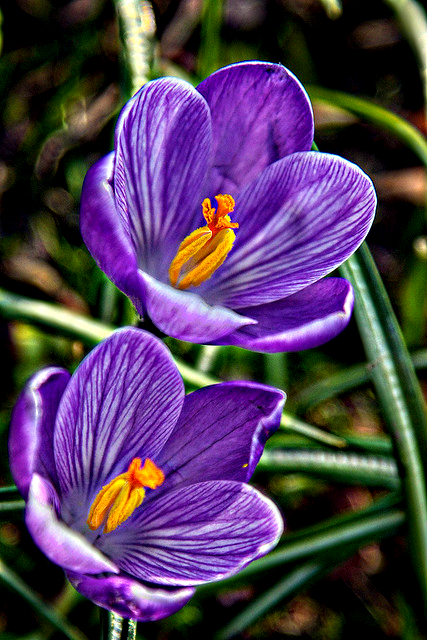}}\hfill
		{\includegraphics[width=0.14\textwidth]{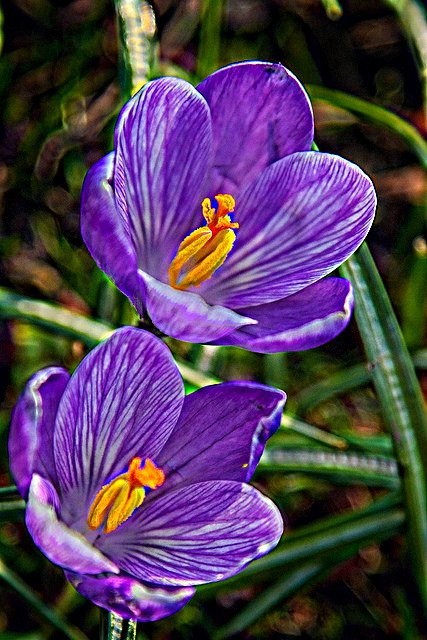}}\hfill
		{\includegraphics[width=0.14\textwidth]{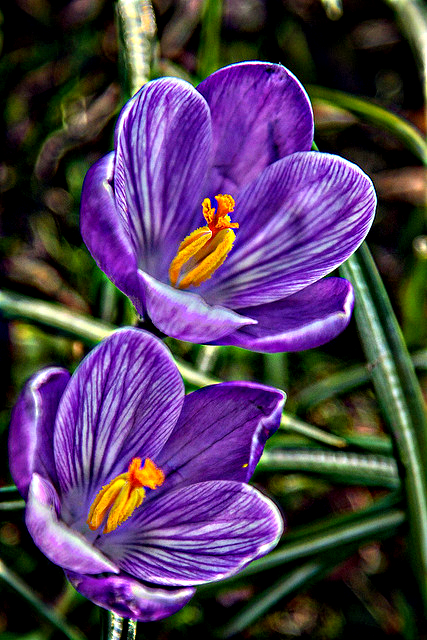}}\hfill
		{\includegraphics[width=0.14\textwidth]{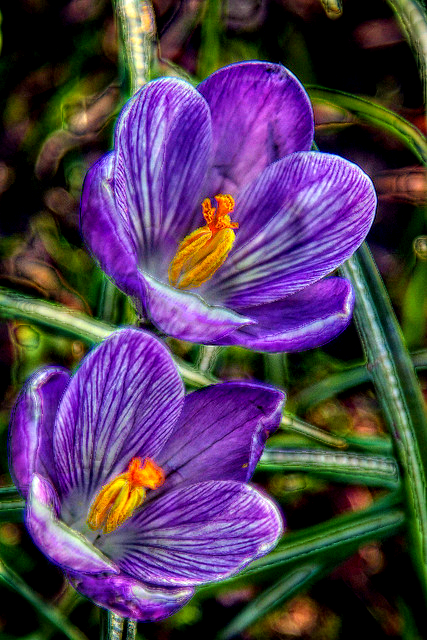}}\hfill
		{\includegraphics[width=0.14\textwidth]{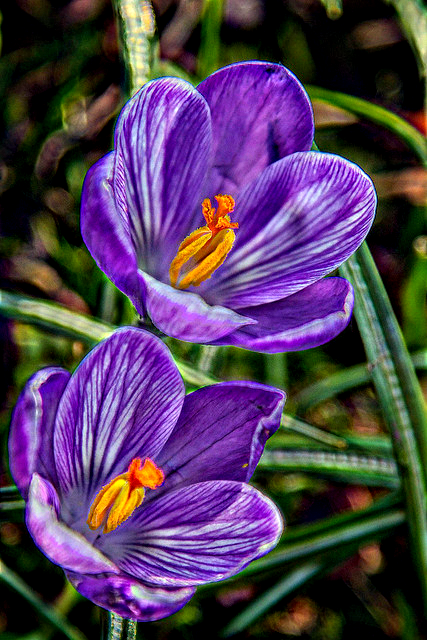}}\hfill
		\subfloat[Input]
		{\includegraphics[width=0.14\textwidth]{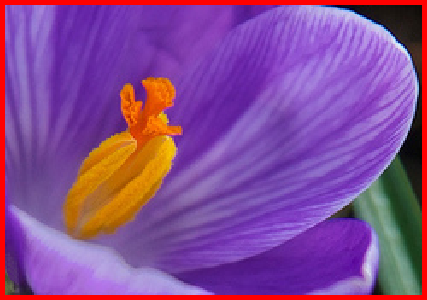}}\hfill
		\subfloat[GF\cite{he2012guided}]
		{\includegraphics[width=0.14\textwidth]{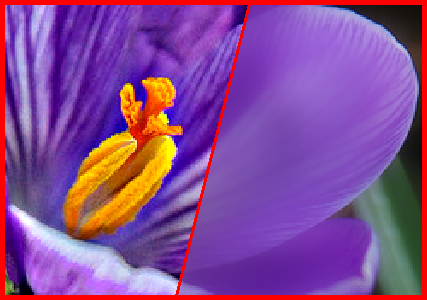}}\hfill
		\subfloat[WLS\cite{farbman2008edge}]
		{\includegraphics[width=0.14\textwidth]{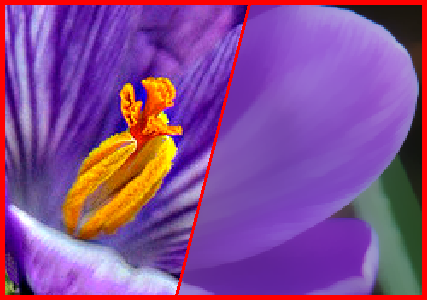}}\hfill
		\subfloat[BM3D\cite{dabov2007image}]
		{\includegraphics[width=0.14\textwidth]{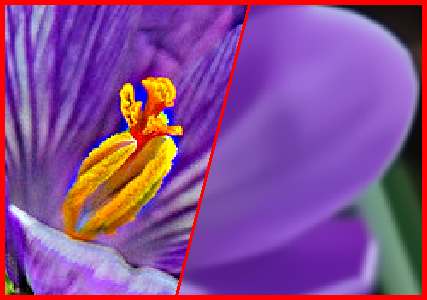}}\hfill
		\subfloat[TGV\cite{knoll2011second}]
		{\includegraphics[width=0.14\textwidth]{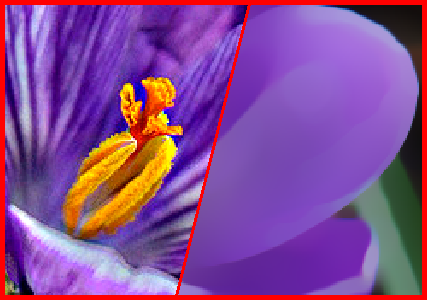}}\hfill
		\subfloat[$L_0$\cite{xu2011image}]
		{\includegraphics[width=0.14\textwidth]{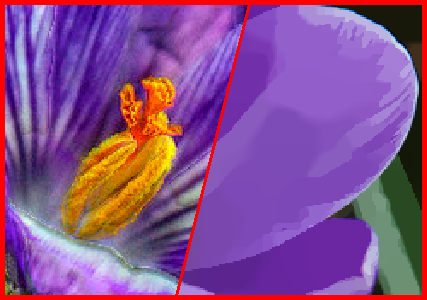}}\hfill
		\subfloat[Ours]
		{\includegraphics[width=0.14\textwidth]{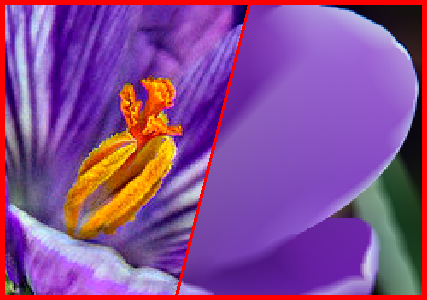}}\hfill
	\end{center}
	\caption{Image enhancement with ($\times$3.5) boosted detail layers. (a) Input image, (b) Guided filter ($w = 15, \epsilon = 0.25^2$)\cite{he2012guided},  (c) WLS filter ($\alpha = 0.5, \lambda = 1.2$)\cite{farbman2008edge},  (d)BM3D\cite{dabov2007image}, (e) TGV ($\lambda \!=\! 0.02, \alpha_0\!=\!0.05, \alpha_1 \!=\! 2e\!-\!3$)\cite{knoll2011second}, (f) $L_0$ Smoothing ($\lambda = 8e\!-\!3$)\cite{xu2011image}, (g) Our result ($\alpha = 1.0, \lambda =0.01$). All filters are configured to produce a similar level smoothing results. MAE: (b)$\scriptsize{\sim}$(g): 0.0319, 0.0321, 0.0321, 0.0319, 0.0320, 0.0321. (Zoom in for better visual comparison).}
\end{figure*} %

\section{Applications}
\label{sec:applications}

Similar to many existing filtering methods, it is possible to use our semi-sparsity model in various signal processing fields. We describe several representative ones involving 2D images, including image detail manipulation, high dynamic range (HDR) image compression, image stylization and abstraction.

\subsection{Details Manipulation:} 

\begin{figure*}[!t]
	\begin{center}
		\subfloat[Input]
		{\includegraphics[width=0.195\textwidth]{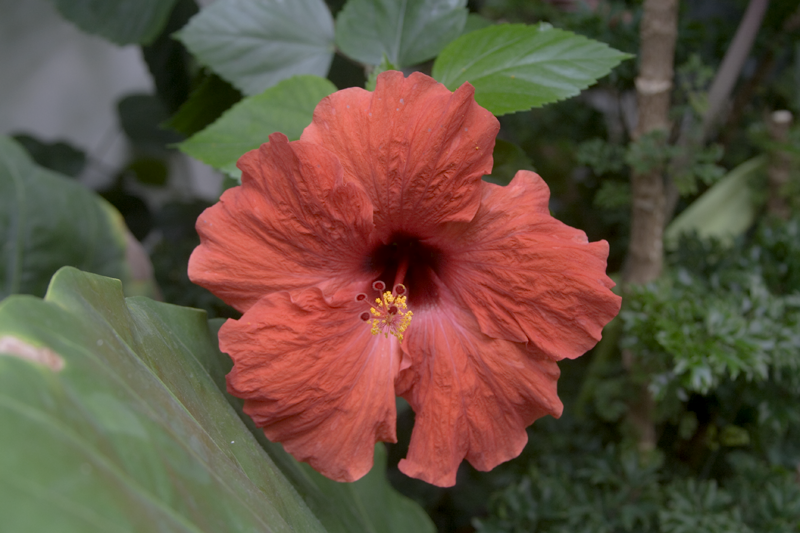}}\hfill
		\subfloat[Small]
		{\includegraphics[width=0.195\textwidth]{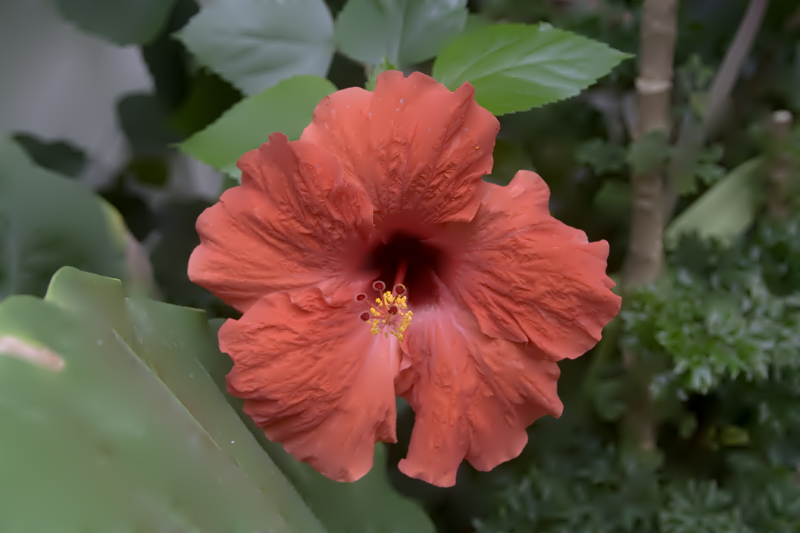}}\hfill
		\subfloat[Median]
		{\includegraphics[width=0.195\textwidth]{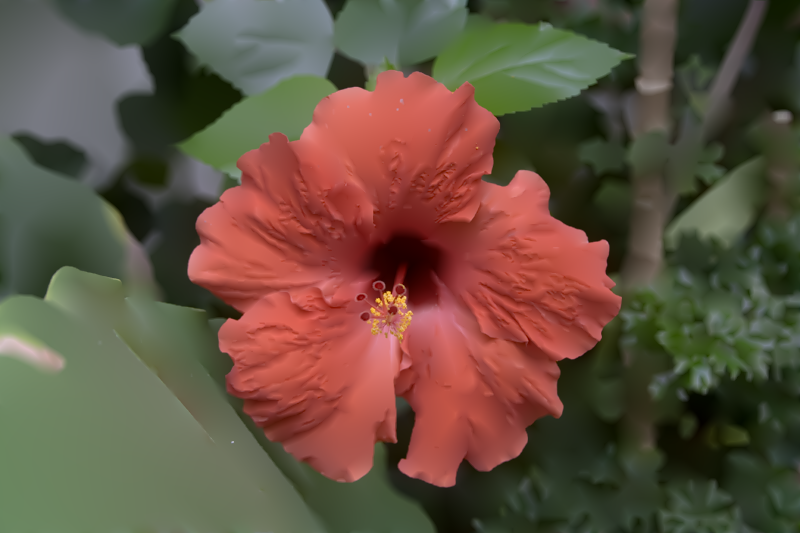}}\hfill
		\subfloat[Coarse]
		{\includegraphics[width=0.195\textwidth]{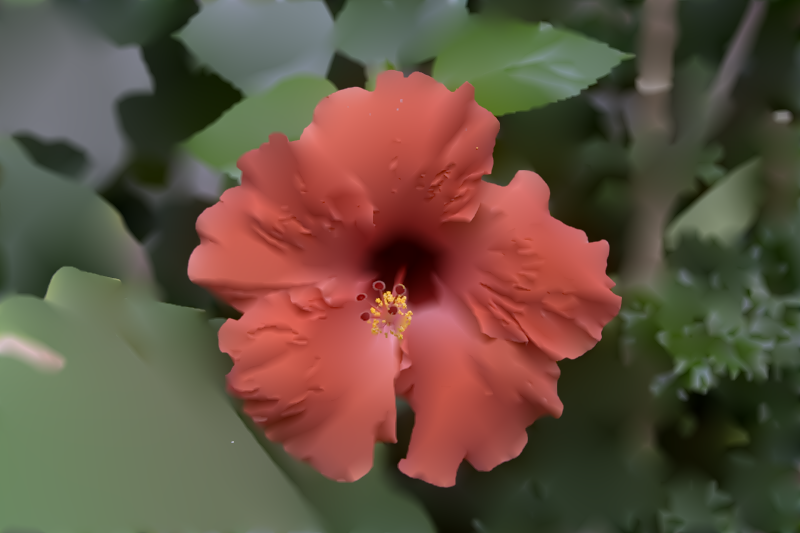}}\hfill
		\subfloat[Merged Result]
		{\includegraphics[width=0.195\textwidth]{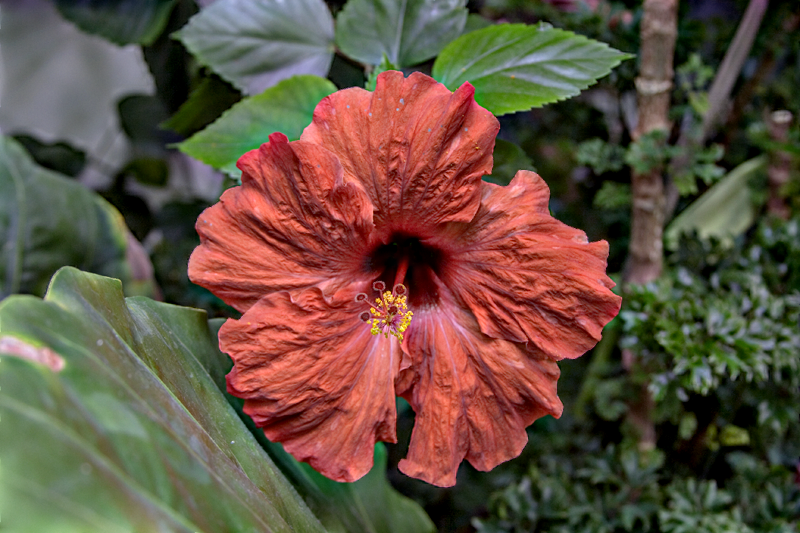}}\hfill
	\end{center}
	\caption{Multi-scale details manipulation. Three scales: (b) small ($\alpha = 0.01, \lambda =0.001$) and (c) median ($\alpha = 0.1, \lambda =0.005$), (d) coarse ($\alpha = 1.0, \lambda =0.02$), and (e) is obtained by scaling ($\times 2$) detail-layers.}
	\label{fig:fig8}
\end{figure*}

In many imaging systems, the captured images may suffer from degradation in details for the inappropriate focus or parameter-settings in image tone mapping, denoising, and so on. In these situations, ``edge-aware'' filtering methods can be used to help recover or exaggerate the degraded details. In general, it is popular to use the smoothing filters to decompose an image into two components: a piece-wise smoothing base layer and a detail layer; then either of them can be remapped and combined for reconstructing better visual results. We compare the results with the cutting-edge methods\cite{farbman2008edge, tomasi1998bilateral, he2012guided, knoll2011second,  xu2011image}. For fairness, all filtering methods produce a similar level of smoothing results. The detail layer is then obtained by subtracting the smoothing base-layer from the original image. The enhanced image is reconstructed by combing the re-scaled ($\times 3.5$) detail layer with the base-layer. As we can see in Fig. 7, the guided filter\cite{he2012guided}, WLS filter\cite{farbman2008edge}, BM3D denoising\cite{dabov2007image} and total generalization variation (TGV)\cite{knoll2011second} methods may cause halo artifacts around strong structures; while $L_0$ smoothing\cite{xu2011image} helps to reduce the halos but leads to crash boundaries due to the inappropriate stair-case filtering results in the polynomial-smoothing regions. In contrast, our semi-sparsity model receives the best result with rich local details but tiny halo artifacts. A multi-scale strategy is also introduced for detail enhancement.  As suggested in WLS filter\cite{farbman2008edge},  the multiple base layers are generated under different smoothing levels, and the detail layers are then obtained by subtracting one from the former smoothing result. As shown in Fig. \ref{fig:fig8}, the small, median and coarse levels are applied for producing smoothing base layers and the final merged result.

\begin{figure*}[!b]
	\begin{center}
		\subfloat[Bilaral Filter\cite{durand2002fast}]
		{\includegraphics[width=0.195\textwidth]{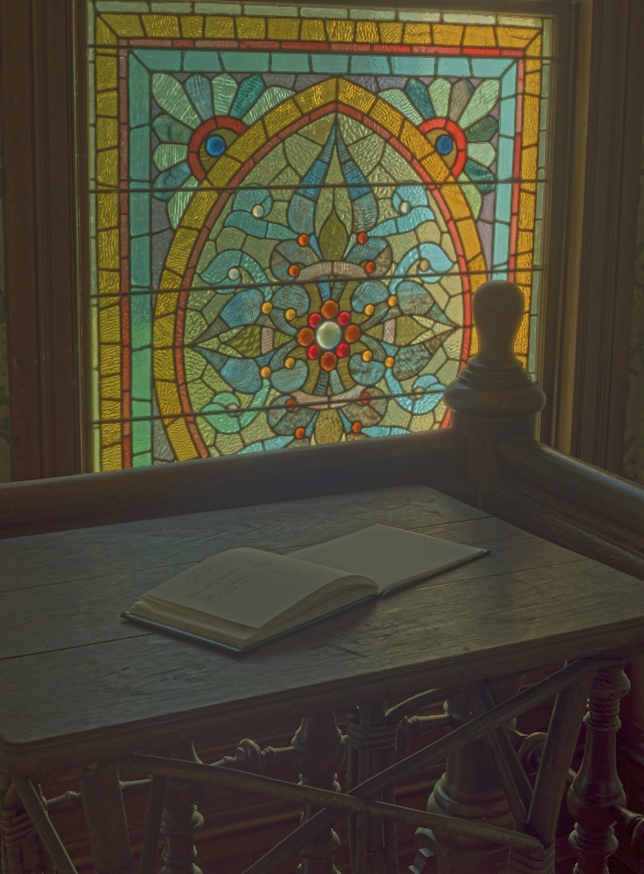}}\hfill
		\subfloat[GD\cite{fattal2002gradient}]
		{\includegraphics[width=0.195\textwidth]{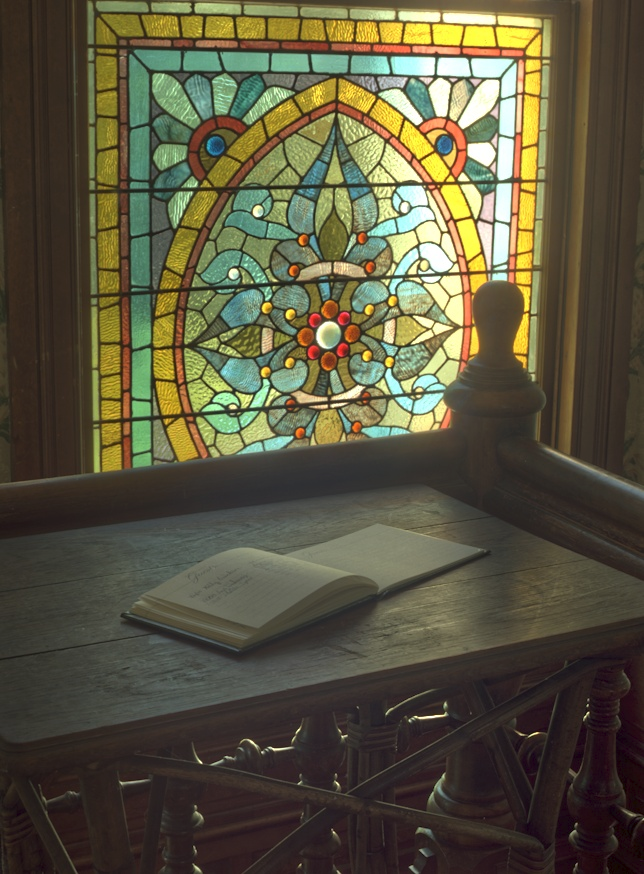}}\hfill
		\subfloat[WLS Filter\cite{farbman2008edge}]
		{\includegraphics[width=0.195\textwidth]{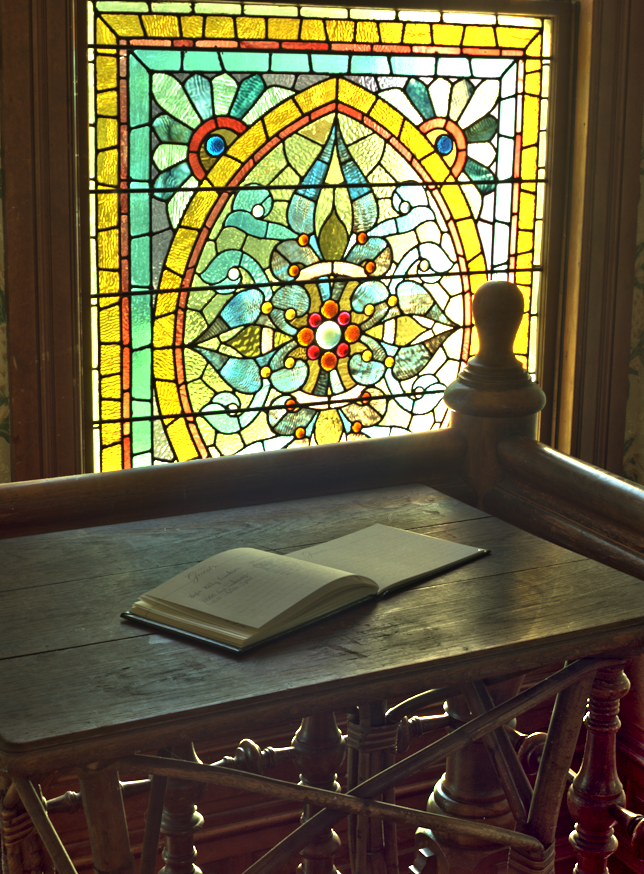}}\hfill
		\subfloat[$L_0$ Smoothing\cite{xu2011image}]
		{\includegraphics[width=0.195\textwidth]{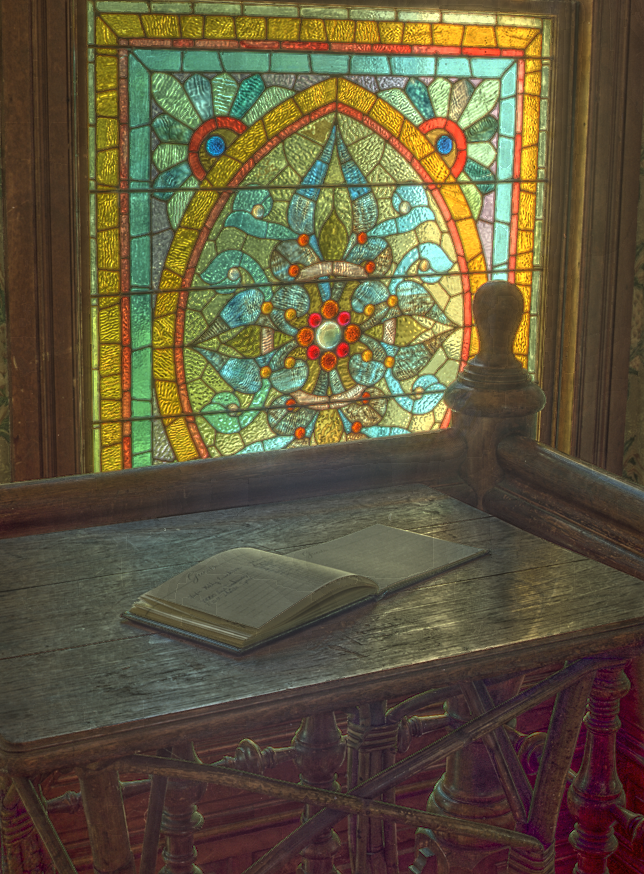}}\hfill
		\subfloat[Our Result ($n=2$)]
		{\includegraphics[width=0.195\textwidth]{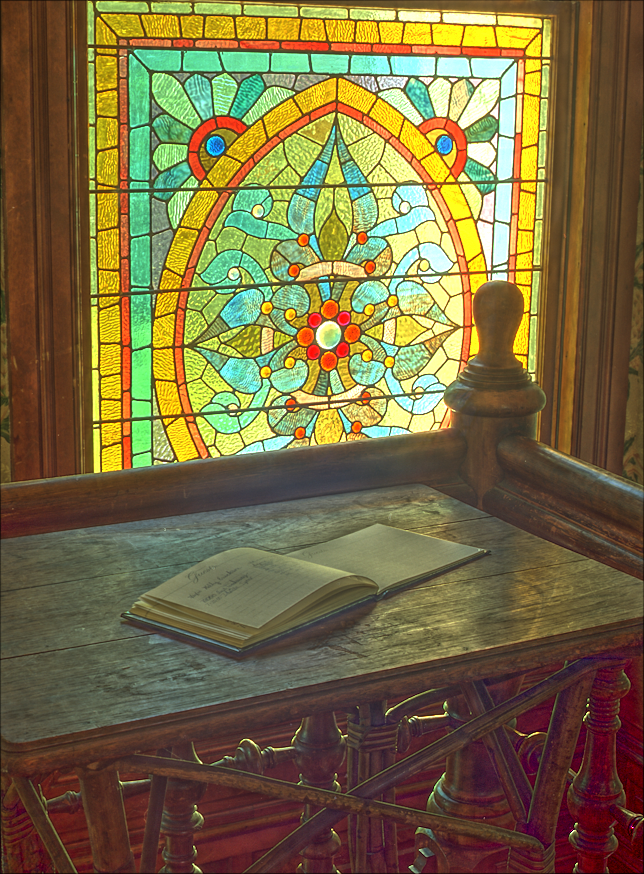}}\hfill
	\end{center}
	\caption{HDR image compression. From left to right: (a) Bilateral filte (BF)\cite{durand2002fast} (b) Gradient domain (GD) compression \cite{fattal2002gradient},  (c) WLS filter,  (d) $L_0$ smoothing and (e) Our result ($\alpha = 0.5, \lambda =0.01)$. The results of (a)$\sim$(d) are adapted from the authors' projects. TMQI (Q): 0.8895, 0.9305, 0.8965, 0.9562.} %
	\label{fig:fig9} 
\end{figure*} %

\subsection{HDR Image Compression:}

This technique arises from the visual-friendly display of high-quality HDR images. It is similar to image details manipulation and an HDR image is usually assumed to be decomposed into a base layer and a detail layer. The base layer is piece-wise smoothing with high dynamic range contrast, which needs to be compressed for display or visual purposes. The compressed smoothing output is then re-combined with the detail layer to reconstruct a new image. The challenge here is to decouple the base layer and maintain reasonable image contrast during dynamic range compression while keeping a balance of the smoothness between the piece-wise areas and sharp discontinuities to avoid halos around strong edges and over-enhancement of spatial local details.

\begin{figure*}[!t]
	\begin{center}
		\subfloat[WLS\cite{farbman2008edge}]
		{\includegraphics[width = 0.195\textwidth, height = 0.23\textheight]{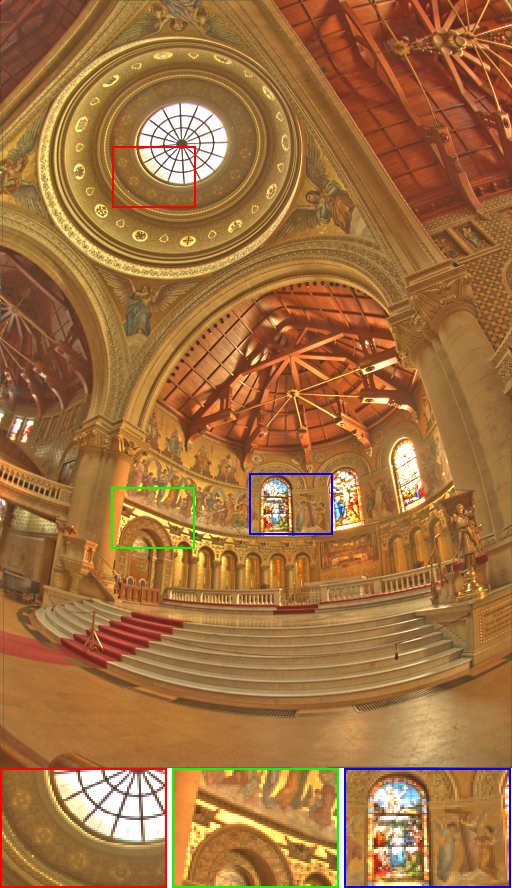}}\hfill
		\subfloat[GF\cite{he2012guided}]
		{\includegraphics[width = 0.195\textwidth, height = 0.23\textheight]{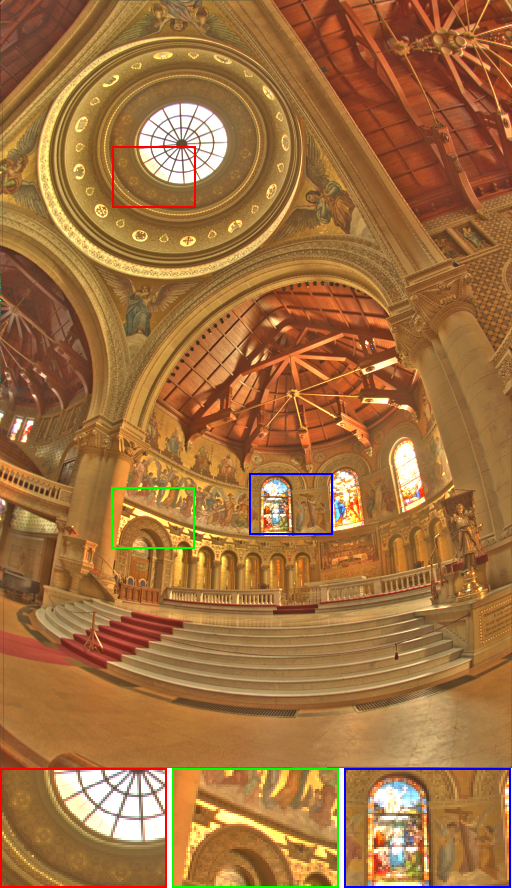}}\hfill
		\subfloat[L0\cite{xu2011image}]
		{\includegraphics[width = 0.195\textwidth, height = 0.23\textheight]{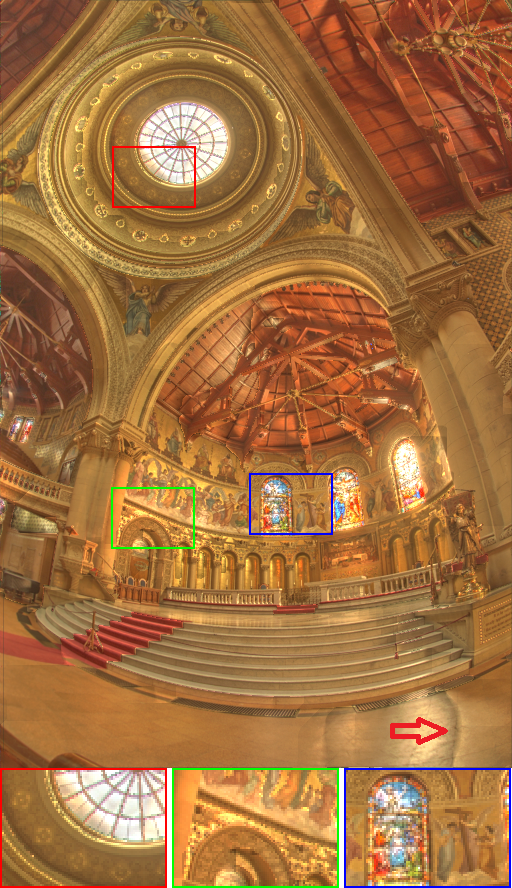}}\hfill
		\subfloat[TGV\cite{knoll2011second}]
		{\includegraphics[width = 0.195\textwidth, height = 0.23\textheight]{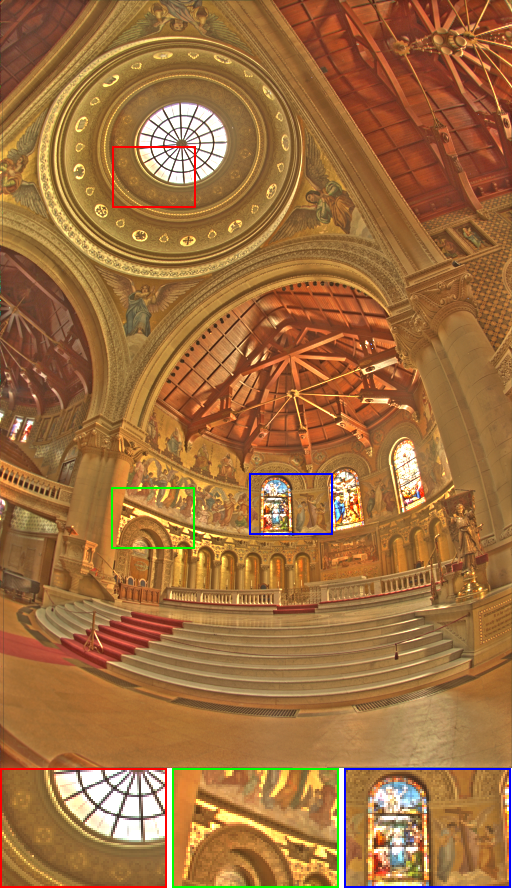}}\hfill
		\subfloat[Ours]
		{\includegraphics[width = 0.195\textwidth, height = 0.23\textheight]{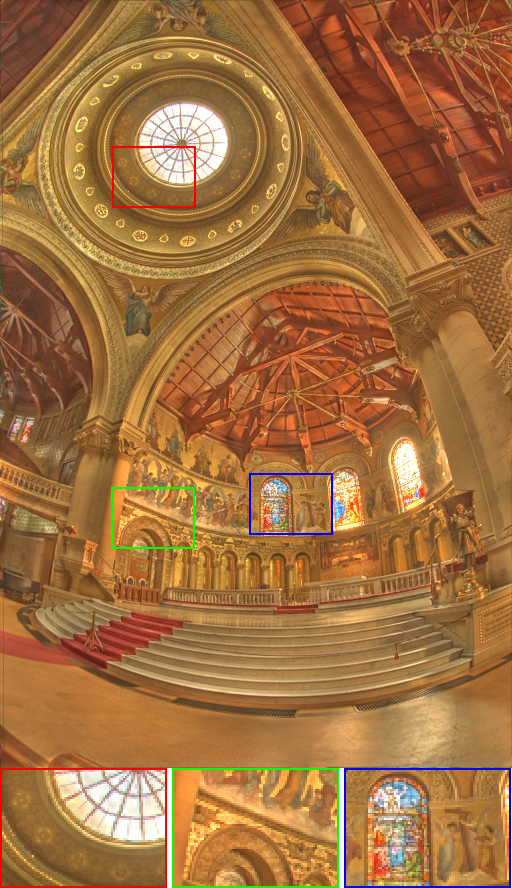}}\hfill
	\end{center}
	\caption{HDR image compression. (a) WLS filter ($\alpha\!=\! 1.2, \lambda\!=\!5.0$)\cite{farbman2008edge}, (b) Guided filter ($r\!=\!15, \delta\!=\![8, 2]$)\cite{he2012guided},  (c) L0 smoothing($\lambda \!=\! 0.5, \kappa \!=\! 1.2$)\cite{xu2011image},  (d) Total generalization variation (TGV) ($\lambda \!=\! 0.02, \alpha_0\!=\!0.1, \alpha_1 \!=\! 0.05$)\cite{knoll2011second}, and (e) Ours ($\alpha \!=\! 1.0, \lambda =0.4$). For fairness, all the filters produce a similar level smoothing results under the same color and saturation configurations.TMQI (Q): 0.9681 0.9624   0.9568 0.9679,  0.9712} 
	\label{fig:fig10}
\end{figure*}

As shown in Fig. \ref{fig:fig9}, bilateral filter\cite{durand2002fast} produces an LDR image with the fine-balance global illumination but the details in local, which is partially recovered by the gradient domain (GD) compression\cite{fattal2002gradient} and WLS filter\cite{farbman2008edge} but they still have limitations in bringing out local tiny details; while $L0$ smoothing\cite{xu2011image} tends to produce exaggerated local details, which is mainly caused by the stair-case fitting results in polynomial smoothing surfaces. In contrast, our semi-sparse filtering method provides a result with appropriate contrast in local and fine-balanced global image illumination.  A similar result is also found in Fig. \ref{fig:fig10}, where we only replace the filter method and the color and saturation share the same configuration as in paper\cite{farbman2008edge} for a fair comparison. Visually, the WLS filter\cite{farbman2008edge}, guided filter\cite{he2012guided}, and total generalization variation (TGV) are possible to produce high-quality LDR results, but they may have limitations in compressing the dynamic range around strong edges; while our method receives a comparable result as the well-known gradient domain\cite{fattal2002gradient} method.  Notice that L0 smoothing\cite{xu2011image} method may also lead to artifacts due to the inappropriate smoothing results in the polynomial-smoothing surfaces.  

To verify the performance of these existing smoothing filters for HDR image compression, we further take a quantitative evaluation based on the Anyhere dataset\cite{greg2003high}. The dataset has 33 widely-used HDR images with different illumination distribution. In general, it is difficut for HDR compression tasks for an objective image quality assessment, because the corresponding ground-truth images are always not available. Recently, no-reference approaches based on statistical models have also shown promising success in predicting the quality of images. As suggested in \cite{huang2021intrinsic}, we use Tone Mapped Image Quality Index (TMQI)\cite{yeganeh2012objective}, Integrated Local Natural Image Quality Evaluator (IL-NIQE)\cite{zhang2015feature} and Neural Image Assessment (NIMA)\cite{talebi2018nima} for evaluation. The TMQI\cite{yeganeh2012objective} index is a full-reference assessment method between HDR image and the output LDR image, in which Structural Fidelity (SF) and Statistical Naturalness (SN) are considered to provide an objective quality assessment. The SF index is based on the multi-scale structural similarity (SSIM) approach\cite{wang2004image} to provide a perceptual predictor of structural fidelity. The SN index is based on a statistic method to takes the natural image statistics of brightness, contrast, visibility and details into account. IL-NIQE\cite{zhang2015feature} and  NIMA\cite{talebi2018nima} are two no-reference image assessments. The former is a learning method based on natural image statistics features derived from color, luminance, gradient and structure information; and the latter attempts to predict consistent aesthetic scores with human opinions using convolutional neural networks. The NIMA\cite{talebi2018nima} method is trained on a large-scale natural image dataset for perceptually-aware no-reference quality assessment. For a fairness comparison, all the filters are configured to produce a similar level smoothing results under the same color and saturation as suggested in \cite{farbman2008edge}. The statistical results are shown in \textbf{Table} \ref{tab:tab1}, where the best two results are highlighted with \textbf{bold} and \textbf{underline}, respectively.  The advantages of the proposed method are also verified with consistent visual-results in Fig. \ref{fig:fig9} and \ref{fig:fig10}. 
\begin{table*}[!t]
	\caption{Quantitative evaluation for HDR image compression on the Anyhere dataset\cite{greg2003high}.}
	\begin{center}
		\setlength{\tabcolsep}{2.5mm}{
			\begin{tabular}{|c|c|c| c| c| c |c |c| c| c| c|}
				\hline
				\multicolumn{2}{|c|}{Metrics} &(b) BF\cite{tomasi1998bilateral}& GF\cite{he2012guided}& TV \cite{rudin1992nonlinear}& WLS \cite{farbman2008edge} & $L_1$-TV\cite{aujol2006structure}& $L_0$ smoothing \cite{xu2011image} & RTV\cite{xu2012structure}   & TGV \cite{knoll2011second} & Ours\\
				\hline
				\multirow{3}{*}{TMQI\cite{yeganeh2012objective}} 	&(SF)    &0.9046    &\underline{0.9059}    &0.9043    &0.9058    &0.9003   & 0.8933    &0.8878    &0.9056     &\textbf{0.9130}\\
										&(SN)    &0.8779    &0.8871    &0.8283    &0.8943    &\underline{0.8967}    &0.8567    &0.8532    &0.8927    &\textbf{0.9254} \\
										&(Q)     &0.9584    &0.9601    &0.9510    &0.9611    &0.9600    &0.9523    &0.9503    &\underline{0.9608} &  \textbf{0.9675}\\ 						  
				\hline
				\hline
				\multicolumn{2}{|c|}{IL-NIQE\cite{zhang2015feature}}      &24.26   &24.79   &23.38  &23.40   &\underline{23.28}  &24.58  &23.90   &23.37&    \textbf{23.02} \\
				\multicolumn{2}{|c|}{NAMA\cite{talebi2018nima}}  	  & 5.43         & 5.48 		&5.27		& \underline{5.52}     	& 5.33        		&  5.21 & 5.30& 5.42& \textbf{5.63}\\
				\hline
		\end{tabular}}
	\end{center}
	\label{tab:tab1}
\end{table*}

\subsection{Image Stylization $\And$ Abstraction: } 

\begin{figure}[!b]
	\centering
	\subfloat[Input]
	{\includegraphics[width=0.16\textwidth]{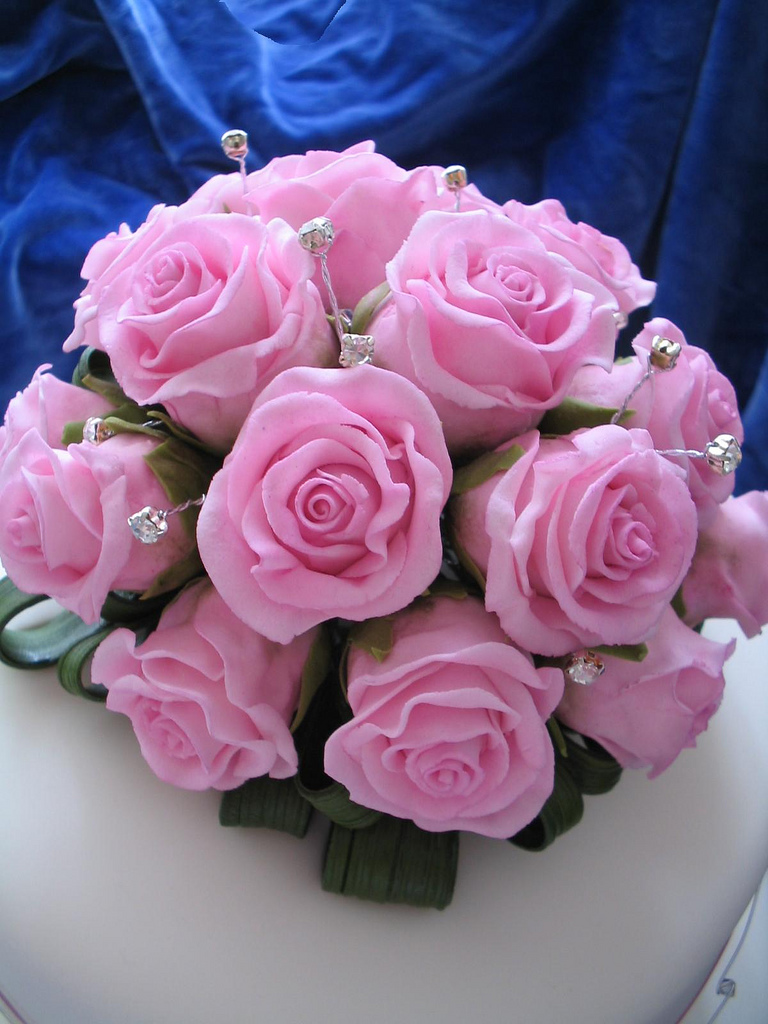}}\hfill
	\subfloat[Pen Drawing]
	{\includegraphics[width=0.16\textwidth]{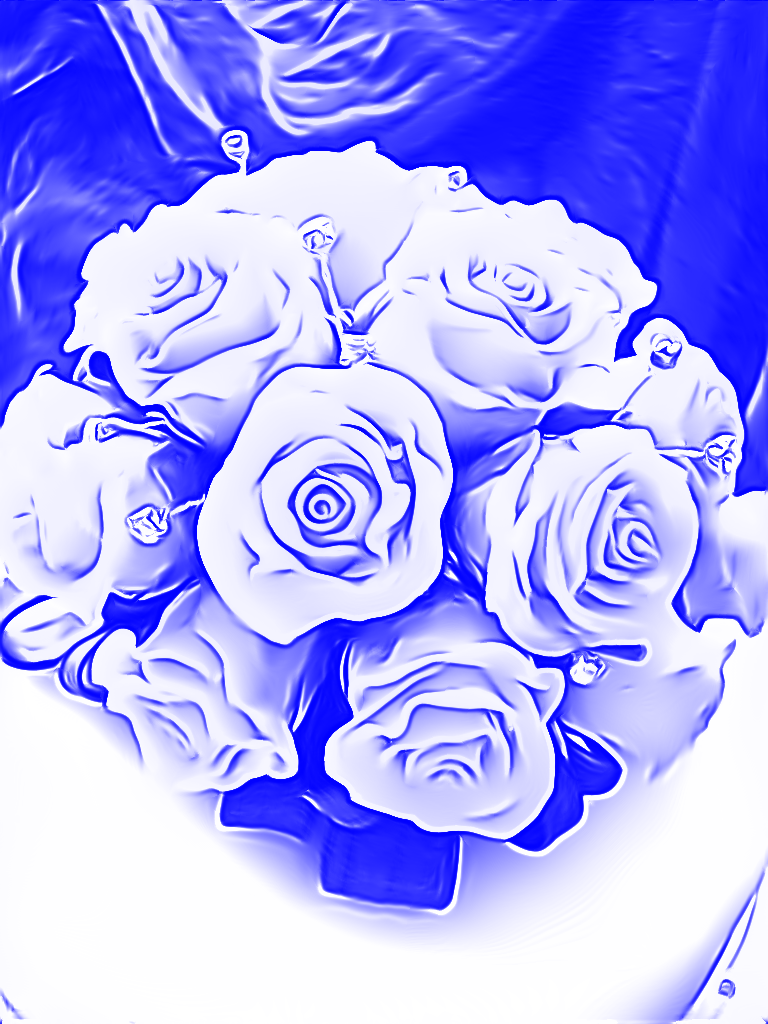}}\hfill
	\subfloat[Color-pencil]
	{\includegraphics[width=0.16\textwidth]{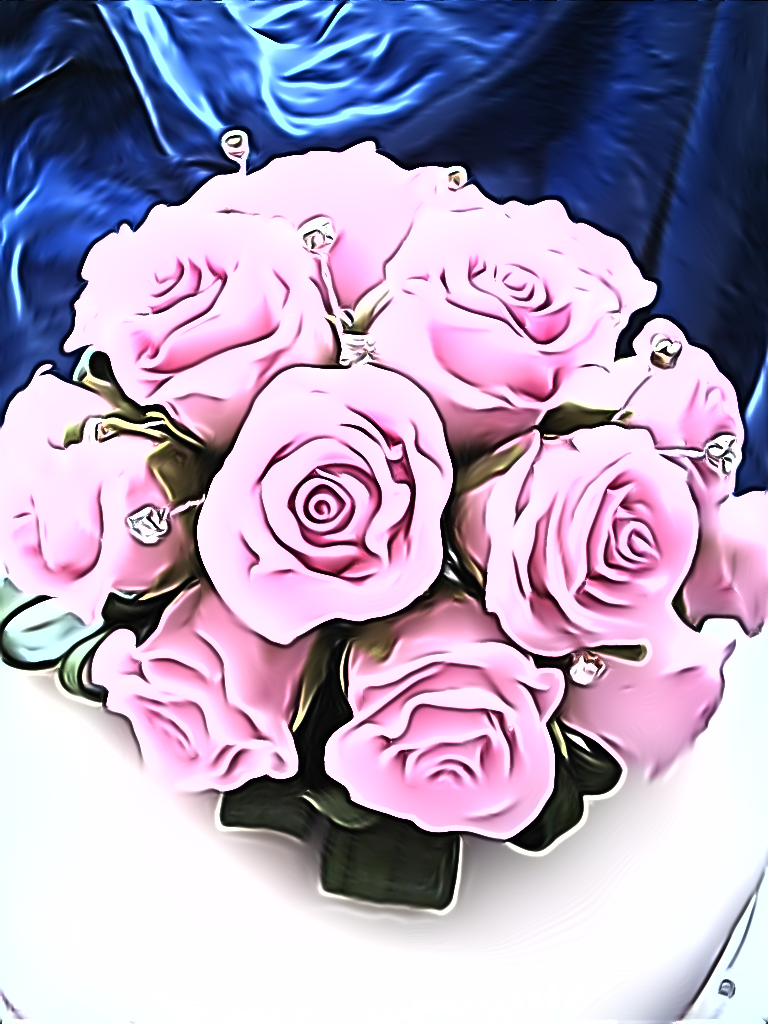}}
	\caption{Image stylization.  From left to right: (a) Input image, (b) Pen drawing, and (c) Color pencil effect.}
	\label{fig:fig11}
\end{figure}

 This interesting task aims to produce different stylized results or non-photorealistic rendering (NPR) of an image. In this case, the proposed semi-sparse filtering method is firstly applied for an image to remove the local textures and details; the main structure of the smoothed image is then, as interpreted in\cite{winnemoller2012xdog}, extracted by using the Difference-of-Gaussian (DoG) operator; and the stylized result is finally attained by merging the filter result and main structures. Due to the simultaneously-smoothing property of the semi-sparse model, it avoids introducing extra edge lines in polynomial surfaces. Despite the isotropic property of the DoG operator, it is usually enough to produce visual-friendly results. As shown in\cite{winnemoller2012xdog}, it is also possible to produce different artistic effects such as hatching, woodcut, watercolor, and so on. We show two simple examples: pen and color-pencil drawing effects in Fig. \ref{fig:fig11}. One can also combine our semi-sparse smoothing filter model with more complex configurations as illustrated in\cite{lu2012combining} to produce more reasonable and aesthetic stylized results.  

\section{Conclusion}
\label{sec:conclusion}

This paper has described a simple semi-sparse minimization scheme for smoothing filters, which is formulated as a general extension of $L_0$-norm minimization in higher-order gradient domains. We demonstrate the virtue of the proposed method in preserving the singularities and fitting the polynomial-smoothing surfaces. We also show its benefits and advantages with a series of experimental results in both image processing and computer vision. One of the drawbacks is the high computational cost of solving the iterative minimization problem, which could be partially alleviated by resorting to more efficient solutions with a faster convergence rate. We leave the work for future studies.

\section*{Acknowledgments}
We thank the anonymous reviewers for their careful reading of our manuscript and insightful comments and suggestions. We also appreciate for the related online resources, including images, codes, software, and so on.


\begin{thebibliography}{1}
	\providecommand{\url}[1]{#1}
	\csname url@samestyle\endcsname
	\providecommand{\newblock}{\relax}
	\providecommand{\bibinfo}[2]{#2}
	\providecommand{\BIBentrySTDinterwordspacing}{\spaceskip=0pt\relax}
	\providecommand{\BIBentryALTinterwordstretchfactor}{4}
	\providecommand{\BIBentryALTinterwordspacing}{\spaceskip=\fontdimen2\font plus
		\BIBentryALTinterwordstretchfactor\fontdimen3\font minus
		\fontdimen4\font\relax}
	\providecommand{\BIBforeignlanguage}[2]{{%
			\expandafter\ifx\csname l@#1\endcsname\relax
			\typeout{** WARNING: IEEEtran.bst: No hyphenation pattern has been}%
			\typeout{** loaded for the language `#1'. Using the pattern for}%
			\typeout{** the default language instead.}%
			\else
			\language=\csname l@#1\endcsname
			\fi
			#2}}
	\providecommand{\BIBdecl}{\relax}
	\BIBdecl
	
	\bibitem{donoho1995noising}
	D.~L. Donoho, ``De-noising by soft-thresholding,'' \emph{IEEE transactions on
		information theory}, vol.~41, no.~3, pp. 613--627, 1995.
	
	\bibitem{perona1990scale}
	P.~Perona and J.~Malik, ``Scale-space and edge detection using anisotropic
	diffusion,'' \emph{IEEE Transactions on pattern analysis and machine
		intelligence}, vol.~12, no.~7, pp. 629--639, 1990.
	
	\bibitem{farbman2008edge}
	Z.~Farbman, R.~Fattal, D.~Lischinski, and R.~Szeliski, ``Edge-preserving
	decompositions for multi-scale tone and detail manipulation,'' in \emph{ACM
		Transactions on Graphics (TOG)}, vol.~27, no.~3.\hskip 1em plus 0.5em minus
	0.4em\relax ACM, 2008, p.~67.
	
	\bibitem{he2012guided}
	K.~He, J.~Sun, and X.~Tang, ``Guided image filtering,'' \emph{IEEE transactions
		on pattern analysis and machine intelligence}, vol.~35, no.~6, pp.
	1397--1409, 2012.
	
	\bibitem{tomasi1998bilateral}
	C.~Tomasi and R.~Manduchi, ``Bilateral filtering for gray and color images.''
	in \emph{Iccv}, vol.~98, no.~1, 1998, p.~2.
	
	\bibitem{paris2011local}
	S.~Paris, S.~W. Hasinoff, and J.~Kautz, ``Local laplacian filters: Edge-aware
	image processing with a laplacian pyramid.'' \emph{ACM Trans. Graph.},
	vol.~30, no.~4, p.~68, 2011.
	
	\bibitem{xu2011image}
	L.~Xu, C.~Lu, Y.~Xu, and J.~Jia, ``Image smoothing via l0 gradient
	minimization,'' in \emph{ACM Transactions on Graphics (TOG)}, vol.~30,
	no.~6.\hskip 1em plus 0.5em minus 0.4em\relax ACM, 2011, p. 174.
	
	\bibitem{winnemoller2012xdog}
	H.~Winnem{\"o}ller, J.~E. Kyprianidis, and S.~C. Olsen, ``Xdog: an extended
	difference-of-gaussians compendium including advanced image stylization,''
	\emph{Computers \& Graphics}, vol.~36, no.~6, pp. 740--753, 2012.
	
	\bibitem{lu2012combining}
	C.~Lu, L.~Xu, and J.~Jia, ``Combining sketch and tone for pencil drawing
	production,'' in \emph{Proceedings of the Symposium on Non-Photorealistic
		Animation and Rendering}.\hskip 1em plus 0.5em minus 0.4em\relax Citeseer,
	2012, pp. 65--73.
	
	\bibitem{weiss2006fast}
	B.~Weiss, ``Fast median and bilateral filtering,'' \emph{Acm Transactions on
		Graphics (TOG)}, vol.~25, no.~3, pp. 519--526, 2006.
	
	\bibitem{fattal2002gradient}
	R.~Fattal, D.~Lischinski, and M.~Werman, ``Gradient domain high dynamic range
	compression,'' in \emph{ACM transactions on graphics (TOG)}, vol.~21,
	no.~3.\hskip 1em plus 0.5em minus 0.4em\relax ACM, 2002, pp. 249--256.
	
	\bibitem{rudin1992nonlinear}
	L.~I. Rudin, S.~Osher, and E.~Fatemi, ``Nonlinear total variation based noise
	removal algorithms,'' \emph{Physica D: nonlinear phenomena}, vol.~60, no.
	1-4, pp. 259--268, 1992.
	
	\bibitem{min2014fast}
	D.~Min, S.~Choi, J.~Lu, B.~Ham, K.~Sohn, and M.~N. Do, ``Fast global image
	smoothing based on weighted least squares,'' \emph{IEEE Transactions on Image
		Processing}, vol.~23, no.~12, pp. 5638--5653, 2014.
	
	\bibitem{ono2017l}
	S.~Ono, ``$ l\_ $\{$0$\}$ $ gradient projection,'' \emph{IEEE Transactions on
		Image Processing}, vol.~26, no.~4, pp. 1554--1564, 2017.
	
	\bibitem{ye2013sparse}
	C.~Ye, D.~Tao, M.~Song, D.~W. Jacobs, and M.~Wu, ``Sparse norm filtering,''
	\emph{arXiv preprint arXiv:1305.3971}, 2013.
	
	\bibitem{milanfar2012tour}
	P.~Milanfar, ``A tour of modern image filtering: New insights and methods, both
	practical and theoretical,'' \emph{IEEE signal processing magazine}, vol.~30,
	no.~1, pp. 106--128, 2012.
	
	\bibitem{buades2011non}
	A.~Buades, B.~Coll, and J.-M. Morel, ``Non-local means denoising,'' \emph{Image
		Processing On Line}, vol.~1, pp. 208--212, 2011.
	
	\bibitem{dabov2007image}
	K.~Dabov, A.~Foi, V.~Katkovnik, and K.~Egiazarian, ``Image denoising by sparse
	3-d transform-domain collaborative filtering,'' \emph{IEEE Transactions on
		image processing}, vol.~16, no.~8, pp. 2080--2095, 2007.
	
	\bibitem{barron2016fast}
	J.~T. Barron and B.~Poole, ``The fast bilateral solver,'' in \emph{European
		conference on computer vision}.\hskip 1em plus 0.5em minus 0.4em\relax
	Springer, 2016, pp. 617--632.
	
	\bibitem{elad2005retinex}
	M.~Elad, ``Retinex by two bilateral filters,'' in \emph{International
		Conference on Scale-Space Theories in Computer Vision}.\hskip 1em plus 0.5em
	minus 0.4em\relax Springer, 2005, pp. 217--229.
	
	\bibitem{gastal2011domain}
	E.~S. Gastal and M.~M. Oliveira, ``Domain transform for edge-aware image and
	video processing,'' in \emph{ACM Transactions on Graphics (ToG)}, vol.~30,
	no.~4.\hskip 1em plus 0.5em minus 0.4em\relax ACM, 2011, p.~69.
	
	\bibitem{li2016fast}
	Y.~Li, D.~Min, M.~N. Do, and J.~Lu, ``Fast guided global interpolation for
	depth and motion,'' in \emph{European Conference on Computer Vision}.\hskip
	1em plus 0.5em minus 0.4em\relax Springer, 2016, pp. 717--733.
	
	\bibitem{milanfar2013}
	P.~Milanfar, ``Symmetrizing smoothing filters,'' \emph{SIAM Journal on Imaging
		Sciences}, vol.~6, no.~1, pp. 263--284, 2013.
	
	\bibitem{takeda2007kernel}
	H.~Takeda, S.~Farsiu, and P.~Milanfar, ``Kernel regression for image processing
	and reconstruction,'' \emph{IEEE Transactions on image processing}, vol.~16,
	no.~2, pp. 349--366, 2007.
	
	\bibitem{grasmair2010anisotropic}
	M.~Grasmair and F.~Lenzen, ``Anisotropic total variation filtering,''
	\emph{Applied Mathematics \& Optimization}, vol.~62, no.~3, pp. 323--339,
	2010.
	
	\bibitem{lefkimmiatis2015structure}
	S.~Lefkimmiatis, A.~Roussos, P.~Maragos, and M.~Unser, ``Structure tensor total
	variation,'' \emph{SIAM Journal on Imaging Sciences}, vol.~8, no.~2, pp.
	1090--1122, 2015.
	
	\bibitem{xu2012structure}
	L.~Xu, Q.~Yan, Y.~Xia, and J.~Jia, ``Structure extraction from texture via
	relative total variation,'' \emph{ACM Transactions on Graphics (TOG)},
	vol.~31, no.~6, p. 139, 2012.
	
	\bibitem{kim2017fast}
	Y.~Kim, D.~Min, B.~Ham, and K.~Sohn, ``Fast domain decomposition for global
	image smoothing,'' \emph{IEEE Transactions on Image Processing}, vol.~26,
	no.~8, pp. 4079--4091, 2017.
	
	\bibitem{nguyen2015fast}
	R.~M. Nguyen and M.~S. Brown, ``Fast and effective {${L}_0$} gradient
	minimization by region fusion,'' in \emph{Proceedings of the IEEE
		international conference on computer vision}, 2015, pp. 208--216.
	
	\bibitem{louchet2011total}
	C.~Louchet and L.~Moisan, ``Total variation as a local filter,'' \emph{SIAM
		Journal on Imaging Sciences}, vol.~4, no.~2, pp. 651--694, 2011.
	
	\bibitem{bredies2010total}
	K.~Bredies, K.~Kunisch, and T.~Pock, ``Total generalized variation,''
	\emph{SIAM Journal on Imaging Sciences}, vol.~3, no.~3, pp. 492--526, 2010.
	
	\bibitem{knoll2011second}
	F.~Knoll, K.~Bredies, T.~Pock, and R.~Stollberger, ``Second order total
	generalized variation (tgv) for mri,'' \emph{Magnetic resonance in medicine},
	vol.~65, no.~2, pp. 480--491, 2011.
	
	\bibitem{parisotto2020higher}
	S.~Parisotto, J.~Lellmann, S.~Masnou, and C.-B. Schonlieb, ``Higher-order total
	directional variation: Imaging applications,'' \emph{SIAM Journal on Imaging
		Sciences}, vol.~13, no.~4, pp. 2063--2104, 2020.
	
	\bibitem{cai2012image}
	J.-F. Cai, B.~Dong, S.~Osher, and Z.~Shen, ``Image restoration: total
	variation, wavelet frames, and beyond,'' \emph{Journal of the American
		Mathematical Society}, vol.~25, no.~4, pp. 1033--1089, 2012.
	
	\bibitem{donoho2003optimally}
	D.~L. Donoho and M.~Elad, ``Optimally sparse representation in general
	(non-orthogonal) dictionaries via {${L}_1$} minimization,'' \emph{Proceedings
		of the National Academy of Sciences}, vol. 100, no.~5, pp. 2197--2202, 2003.
	
	\bibitem{elad2010sparse}
	M.~Elad, \emph{Sparse and redundant representations: from theory to
		applications in signal and image processing}.\hskip 1em plus 0.5em minus
	0.4em\relax Springer Science \& Business Media, 2010.
	
	\bibitem{donoho2006compressed}
	D.~L. Donoho, ``Compressed sensing,'' \emph{IEEE Transactions on information
		theory}, vol.~52, no.~4, pp. 1289--1306, 2006.
	
	\bibitem{gong2014image}
	Y.~Gong and I.~F. Sbalzarini, ``Image enhancement by gradient distribution
	specification,'' in \emph{Asian Conference on Computer Vision}.\hskip 1em
	plus 0.5em minus 0.4em\relax Springer, 2014, pp. 47--62.
	
	\bibitem{kodak}
	``{Kodak Lossless True Color Image Suite,}'' 2013. [Online] Available: 
	\url{http://r0k.us/graphics/kodak}, accessed: 2021-09-30.
	
	\bibitem{aujol2006structure}
	J.-F. Aujol, G.~Gilboa, T.~Chan, and S.~Osher, ``Structure-texture image
	decomposition—modeling, algorithms, and parameter selection,''
	\emph{International journal of computer vision}, vol.~67, no.~1, pp.
	111--136, 2006.
	
	\bibitem{tropp2004greed}
	J.~A. Tropp, ``Greed is good: Algorithmic results for sparse approximation,''
	\emph{IEEE Transactions on Information theory}, vol.~50, no.~10, pp.
	2231--2242, 2004.
	
	\bibitem{chen2001atomic}
	S.~S. Chen, D.~L. Donoho, and M.~A. Saunders, ``Atomic decomposition by basis
	pursuit,'' \emph{SIAM review}, vol.~43, no.~1, pp. 129--159, 2001.
	
	\bibitem{wang2008new}
	Y.~Wang, J.~Yang, W.~Yin, and Y.~Zhang, ``A new alternating minimization
	algorithm for total variation image reconstruction,'' \emph{SIAM Journal on
		Imaging Sciences}, vol.~1, no.~3, pp. 248--272, 2008.
	
	\bibitem{allain2006global}
	M.~Allain, J.~Idier, and Y.~Goussard, ``On global and local convergence of
	half-quadratic algorithms,'' \emph{IEEE Transactions on Image Processing},
	vol.~15, no.~5, pp. 1130--1142, 2006.
	
	\bibitem{blumensath2009iterative}
	T.~Blumensath and M.~E. Davies, ``Iterative hard thresholding for compressed
	sensing,'' \emph{Applied and computational harmonic analysis}, vol.~27,
	no.~3, pp. 265--274, 2009.
	
	\bibitem{sun2017convergence}
	T.~Sun and L.~Cheng, ``Convergence of iterative hard-thresholding algorithm
	with continuation,'' \emph{Optimization Letters}, vol.~11, no.~4, pp.
	801--815, 2017.
	
	\bibitem{boyd2011distributed}
	S.~Boyd, N.~Parikh, and E.~Chu, \emph{Distributed optimization and statistical
		learning via the alternating direction method of multipliers}.\hskip 1em plus
	0.5em minus 0.4em\relax Now Publishers Inc, 2011.
	
	\bibitem{saad2003iterative}
	Y.~Saad, \emph{Iterative methods for sparse linear systems}.\hskip 1em plus
	0.5em minus 0.4em\relax SIAM, 2003.
	
	\bibitem{attouch2013convergence}
	H.~Attouch, J.~Bolte, and B.~F. Svaiter, ``Convergence of descent methods for
	semi-algebraic and tame problems: proximal algorithms, forward--backward
	splitting, and regularized gauss--seidel methods,'' \emph{Mathematical
		Programming}, vol. 137, no.~1, pp. 91--129, 2013.
	
	\bibitem{zeng2016sparse}
	J.~Zeng, S.~Lin, and Z.~Xu, ``Sparse regularization: Convergence of iterative
	jumping thresholding algorithm,'' \emph{IEEE Transactions on Signal
		Processing}, vol.~64, no.~19, pp. 5106--5118, 2016.
	
	\bibitem{he2013mesh}
	L.~He and S.~Schaefer, ``Mesh denoising via l 0 minimization,'' \emph{ACM
		Transactions on Graphics (TOG)}, vol.~32, no.~4, p.~64, 2013.
	
	\bibitem{durand2002fast}
	F.~Durand and J.~Dorsey, ``Fast bilateral filtering for the display of
	high-dynamic-range images,'' in \emph{ACM transactions on graphics (TOG)},
	vol.~21, no.~3.\hskip 1em plus 0.5em minus 0.4em\relax ACM, 2002, pp.
	257--266.
	
	\bibitem{greg2003high}
	\BIBentryALTinterwordspacing
	A.~S. Greg~Ward, ``High dynamic range image examples,'' 2003. [Online].
	Available: \url{https://tex.stackexchange.com}, accessed: 2021-09-30.
	\BIBentrySTDinterwordspacing
	
	\bibitem{huang2021intrinsic}
	J.~Huang, M.~Ruzhansky, Q.~Zhang, and H.~Wang, ``Intrinsic image transfer for
	illumination manipulation,'' \emph{IEEE Transactions on Pattern Analysis and 
	Machine Intelligence}, 2022.
	
	\bibitem{yeganeh2012objective}
	H.~Yeganeh and Z.~Wang, ``Objective quality assessment of tone-mapped images,''
	\emph{IEEE Transactions on Image processing}, vol.~22, no.~2, pp. 657--667,
	2012.
	
	\bibitem{zhang2015feature}
	L.~Zhang, L.~Zhang, and A.~C. Bovik, ``A feature-enriched completely blind
	image quality evaluator,'' \emph{IEEE Transactions on Image Processing},
	vol.~24, no.~8, pp. 2579--2591, 2015.
	
	\bibitem{talebi2018nima}
	H.~Talebi and P.~Milanfar, ``Nima: Neural image assessment,'' \emph{IEEE
		transactions on image processing}, vol.~27, no.~8, pp. 3998--4011, 2018.
	
	\bibitem{wang2004image}
	Z.~Wang, A.~C. Bovik, H.~R. Sheikh, and E.~P. Simoncelli, ``Image quality
	assessment: from error visibility to structural similarity,'' \emph{IEEE
		transactions on image processing}, vol.~13, no.~4, pp. 600--612, 2004.
	
\end{thebibliography}

%

\begin{IEEEbiography}[{\includegraphics[width=1in,height=1.25in,clip,keepaspectratio]{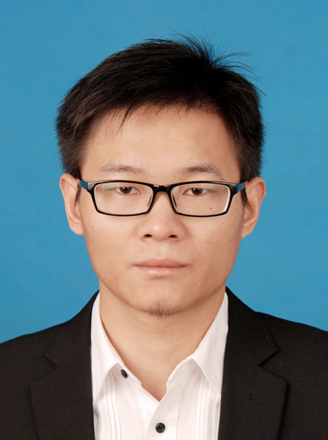}}]{Junqing Huang}
	received the BS degree in Automation from the School of Electrical Engineering, Zhengzhou University, Zhengzhou, China, in 2011, and the MS degree in Mathematics from the School of Mathematical Sciences, Beihang University (BUAA), Beijing, China, in 2015. He is currently a Ph.D. candidate of Department of Mathematics: Analysis, Logic and Discrete Mathematics, Ghent University, Belgium. His research interests include deep learning, image processing, optimal transport and optimization.
\end{IEEEbiography}

\begin{IEEEbiography}[{\includegraphics[width=1in,height=1.25in,clip,keepaspectratio]{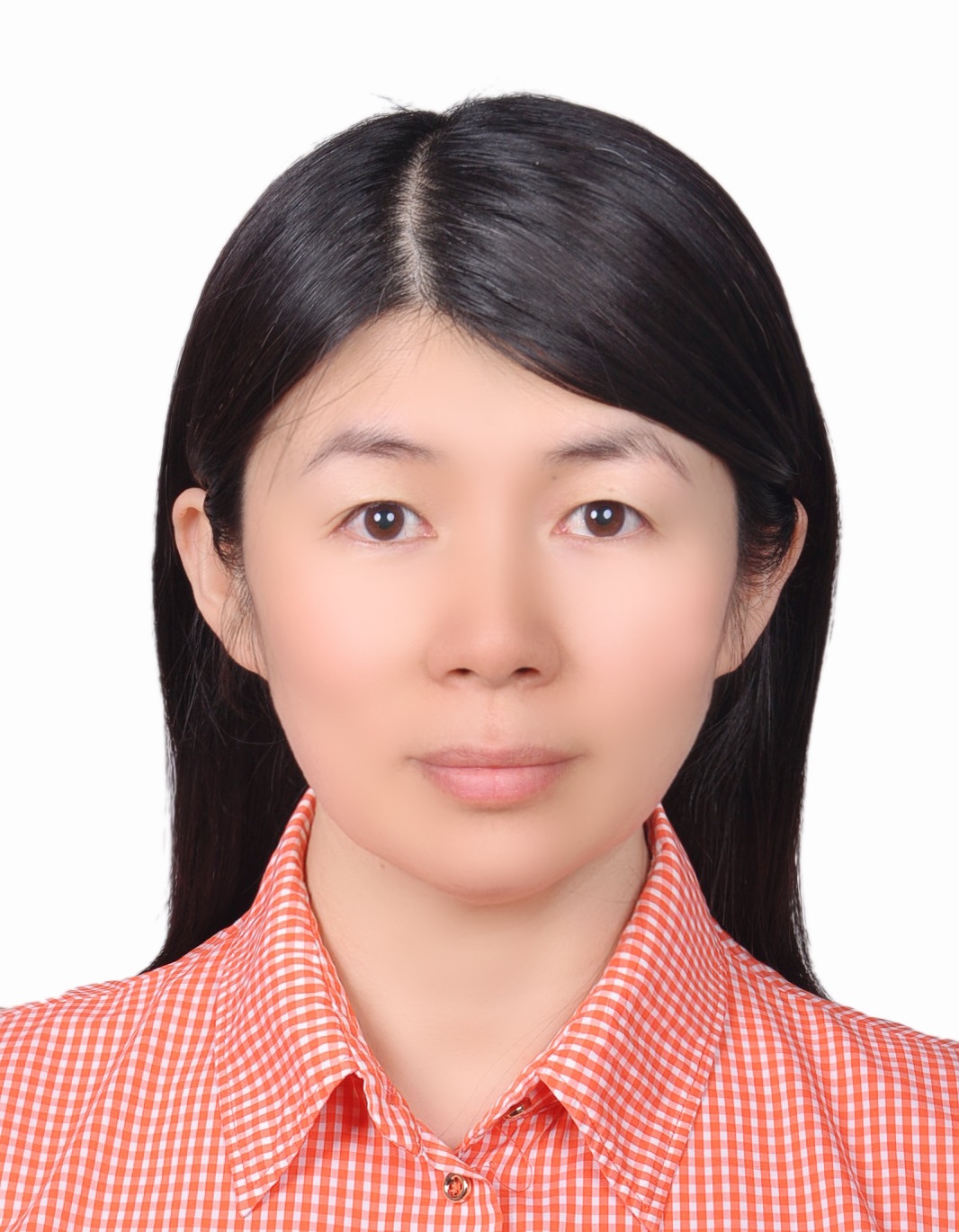}}]{Haihui Wang}	
	received the PhD degree in mathematics from Beijing University, Beijing, China, in 2003. She is currently a full professor in the school of Mathematics and Sciences, Beihang University (BUAA), Beijing, China. Her research interests include artificial intelligence, machine learning, signal and image processing, wavelet analysis and applications.
\end{IEEEbiography}

\begin{IEEEbiography}[{\includegraphics[width=1in,height=1.25in,clip,keepaspectratio]{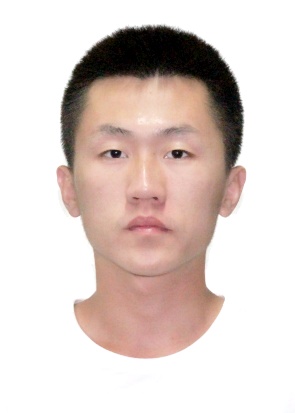}}]{Xuechao Wang}
	received the BS degree in mathematics from Shandong Normal University, Jinan, China, in 2018, and the MS degree in mathematics from Beihang University (BUAA), Beijing, China, in 2021. His research interests include machine and deep learning, computer vision.
\end{IEEEbiography}

\begin{IEEEbiography}[{\includegraphics[width=1in,height=1.25in,clip,keepaspectratio]{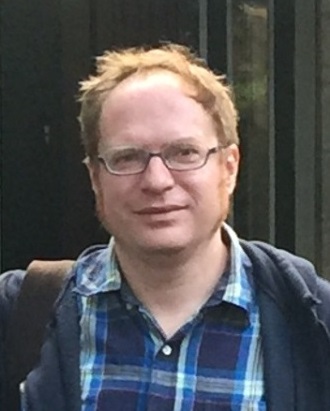}}]{Michael Ruzhansky}
	is currently a senior full professor in Department of Mathematics and a
	Professorship in Special Research Fund (BOF) at Ghent University of Belgium, a Professorship in School of Mathematical Sciences at Queen Mary University of London, UK, and Honorary Professorship in Department of Mathematics at Imperial College London, UK.
	He was awarded by FWO (Belgium) the prestigious Odysseus 1 Project in 2018, he was recipient of several Prizes and Awards: ISAAC Award in 2007,
	Daiwa Adrian Prize in 2010 and Ferran Sunyer I Balaguer Prizes in 2014 and 2018. His research interests include different areas of analysis, in particular, theory of PDEs, microlocal analysis, and harmonic analysis.
\end{IEEEbiography}
\vfill

\end{document}